\documentclass[12pt]{article}

\usepackage[ruled,vlined]{algorithm2e}
\usepackage{microtype}
\usepackage{graphicx}
\usepackage{caption}
\usepackage{subcaption}
\usepackage{booktabs} 
\usepackage{wrapfig}
\usepackage{float}
\usepackage{xspace}
\usepackage{amsfonts}
\usepackage{amsmath}

\usepackage{hyperref}

\newcommand{\Ghat}		{{\widehat{G}}}

\newcommand{\WWLAlgSimp}	{{\sf MorseGraphRecon\xspace}}

\newcommand{\MorseTrain}	{{\sf MorseLabelTrain\xspace}}
\newcommand{\SkeTrain}	{{\sf SkeletonLabelTrain\xspace}}
\newcommand{\Hdis}{\hmm{a}verage Hausdorff distance\xspace}
\newcommand\hmm[1]{\ifnum\ifhmode\spacefactor\else2000\fi>1000 \uppercase{#1}\else#1\fi}

\newcommand{\graphReconAlg}{discrete-Morse based graph reconstruction algorithm}

\newtheorem{theorem}{Theorem}[section]
\newtheorem{definition}[theorem]{Definition}

\title{Road Network Reconstruction from satellite images with  Machine Learning Supported by Topological Methods}
\author{Tamal K. Dey\thanks{Department of Computer Science and Engineering, The Ohio State University. \texttt{tamaldey, yusu@cse.ohio-state.edu, wang.6195@osu.edu}}, Jiayuan Wang\footnotemark[1], Yusu Wang\footnotemark[1]}

\date{}
\begin{document}
\maketitle
\begin{abstract}
Automatic Extraction of road network from satellite images is a goal that can benefit and even enable new technologies. 
Methods that combine machine learning (ML) and computer vision have been proposed in recent years which make the task semi-automatic by requiring the user to provide curated training samples.
The process can be fully automatized if training samples can be produced algorithmically. Of course, this requires a robust algorithm that can reconstruct the road networks from satellite images  reliably so that the output can be fed as training samples. In this work, we develop such a technique by infusing a persistence-guided discrete Morse based graph reconstruction algorithm into ML framework. 

We elucidate our contributions in two phases.
First, in a semi-automatic framework, we combine a \graphReconAlg{} with an existing CNN framework to segment input satellite images. We show that this leads to reconstructions with better connectivity and less noise. Next, in a fully automatic framework, we leverage the power of the \graphReconAlg{} to train a CNN from a collection of images {\bf without labelled data} and use the same algorithm to produce the final output from the segmented images created by the trained CNN. We apply the \graphReconAlg{} iteratively to improve the accuracy of the CNN. We show promising experimental results of this new framework on datasets from SpaceNet Challenge.
\end{abstract}



\section{Introduction}
Layout of road networks is essential for diverse applications in geographic information systems. 
Efficient reconstruction from images and timely updates of road networks are important both for map designs and handling events such as natural disasters. The availability of high-resolution satellite images has enabled such technology in recent years though the process is not fully automatic. 
Currently the road extraction from satellite images is mainly completed manually \cite{van2018spacenet}. 
Doing so automatically or even semi-automatically in a reliable manner is challenging as there are a variety of different types of roads whose images are cluttered with noise and occlusions (by cars/trees etc).

Extracting lane-related information from high resolution satellite images has been addressed in recent years~\cite{gu2015fusion,sun2018combining,zang2017lane}.
Specifically for road extraction, a range of methods that combine machine learning and computer vision methods have been proposed to reconstruct roads using labelled data. These are semi-automatic in the sense that they use manually curated samples to train the classifier.
These methods often consist of two main stages. 
The first stage consists of the background segmentation and the second stage consists of the centerline extraction.
The background segmentation is usually done via machine learning methods 
such as performing feature extraction and  pixel-wise label predictions with SVM \cite{das2011use,shi2014integrated} or CNN \cite{ciresan2012deep}.
More recently, a CNN framework called U-Net \cite{ronneberger2015u} is proposed that outputs the segmentations directly,  
improving the predictions and the running time significantly.
The baseline algorithms for SpaceNet Challenge \cite{van2018spacenet} use the architecture such as U-Net  and PSPNet \cite{zhao2017pyramid}. 
For the second stage, methods like skeleton or medial axis extraction with pre- and post-processing are often used to obtain the final road networks.
However, recovering the correct connections and junctions of roads still remain challenging.
This problem is critical since the road network is often used in routing and false breaks in the extraction lead to unacceptable results. This two-stage approach
can potentially be fully automatized if training samples can be produced algorithmically. 

To achieve the full automatization, one needs to have a direct reconstruction from images that may not be completely faithful but reliable enough to serve as the generator of good training samples. Then, one can iteratively use the technique to improve upon the training samples. This is what we achieve in this work for automatic road reconstruction. It turns out that our direct reconstruction method even improves over the state-of-the-art techniques for semi-automatic reconstructions by providing a more robust algorithm for the second stage. Our direct reconstruction method is a topology-based graph reconstruction algorithm. 
It uses the recent techniques of topological 
persistence~\cite{EH10} and discrete Morse theory~\cite{forman} in topological data analysis. 
This topology-based approach for recovering hidden structures has been proposed and studied recently \cite{DRS15,GDN07,RWS11,weiss2013primal}.
It has been applied to extracting graph-like structures from simulated dark matter density fields \cite{2011MNRAS} and reconstructing road networks from GPS traces \cite{wang2015efficient, dey2017improved}. 
This discrete-Morse based graph reconstruction framework is clean both conceptually and implementation-wise. Most importantly, as it uses a global topological structure to make decisions (instead of using purely local information to decide whether a point is on or off the road), the algorithm is robust to noise, non-uniform sampling of the data, and reliable at recovering junctions. 
Very recently, this graph reconstruction algorithm has been further simplified, and theoretical guarantees of this graph reconstruction algorithm for the case when the signal prevails noise have been provided  \cite{dey2018graph}.


\paragraph{Specific contributions:}
Our contribution is twofold.\\

(1) First, in a semi-automatic framework, we apply the discrete-Morse based graph reconstruction algorithm on the segmented satellite images obtained by a CNN. This, of course, requires user provided training samples to train the CNN.
We show that this leads to reconstructions with better network connectivity and less noise compared to some existing state-of-the-art technique.

(2) More importantly, next, in a fully automatic framework, we develop a novel method to leverage the power of the discrete-Morse based graph reconstruction algorithm to train a CNN from a collection of images {\bf without labelled data} so that it can produce segmentation for new images. 
To elaborate, we start with running the graph reconstruction algorithm on the raw satellite images to obtain some initial reconstructions. 
We then put the pixels from reliable branches of the output graph as positive and others as negative to create the labels for the training, and produce an intermediate CNN classifier.
We predict the segmented images for the training set using this intermediate CNN and then repeat the same process on the output to gradually improve the CNN. 
Our experiment shows that after several iterations of training, the labels computed from the graph reconstruction algorithm become less noisy and the performance of the classifier improves significantly.
If we relax the condition slightly and assume that we know the labels for only 10\% of the train set, we can incorporate this partially labelled data into our framework, and the performance of the classifier becomes even better. 


We experiment on datasets from SpaceNet Challenge \cite{van2018spacenet} which consists of high resolution images for four cities.
For the semi-automatic framework, we compare our results with the results of the winner's algorithm, using the APLS score (defined in  \cite{van2018spacenet}) as well as another metric which we call \Hdis, to evaluate the quality of the reconstructed networks compared to the ground-truth (provided by SpaceNet Challenge).
Overall, our reconstructions tend to have better connectivity and are less noisy.
For the fully automatic framework, we show that the reconstruction quality is significantly improved through our iterative training process. 
Furthermore, our framework can be modified to include a small set of labelled data 
and the accuracy improves as we use more and more labelled data. 

This paper is organized as follows, Section \ref{sec:preliminary} briefly describes the idea of the discrete-Morse based graph reconstruction algorithm. 
Section \ref{sec:aprch}  introduces our semi-automatic framework and fully automatic framework.
Section \ref{sec:exp} provides various experiment results for both frameworks and discusses limitations and future works.

\section{Discrete-Morse based graph reconstruction}
\label{sec:preliminary}

\begin{figure}[t]
\includegraphics[width=\textwidth]{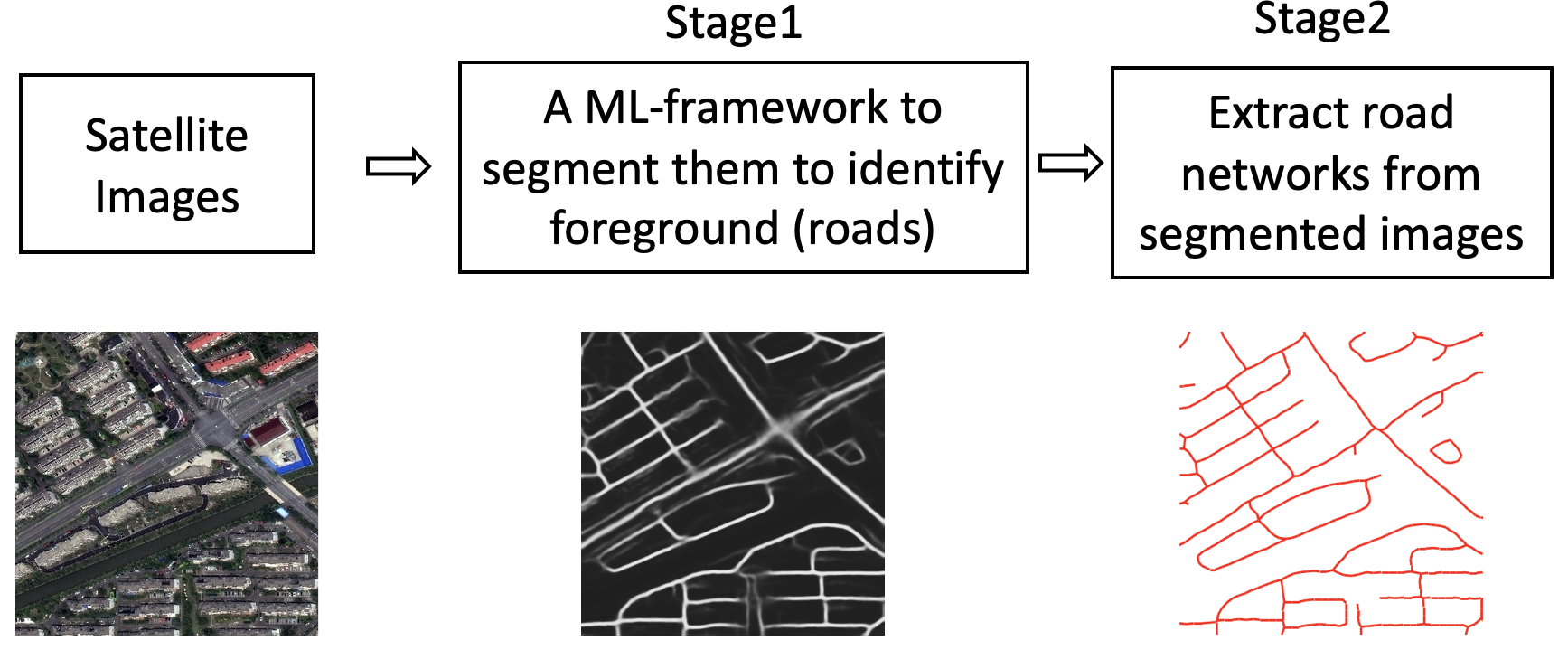}
\caption{High level road network reconstruction from satellite images framework pipeline.
\label{fig:pipeline}}
\end{figure}

On the high level, the road network reconstruction from satellite images framework has two stages; see Figure \ref{fig:pipeline}. 
In Stage 1, we use some machine learning techniques to convert a given satellite image into a segmented image where roughly speaking, the value at each pixel represents the likelihood of this pixel being on / around roads. 
In Stage 2, we extract the hidden road-network (graph) from this segmented image. 

We apply the simplified version of our discrete-Morse based graph reconstruction algorithm \cite{dey2018graph} to extract road networks from the segmented images (sometimes called ``road masking'' in the literature). 
Given that our approach mostly uses this reconstruction algorithm as a black box, we only provide a high-level description of the main ideas here. Interested readers should see \cite{dey2018graph} for more details. 

\begin{figure}[thbp]
    \centering
    \begin{subfigure}[b]{0.4\textwidth}
        \includegraphics[width=\textwidth]{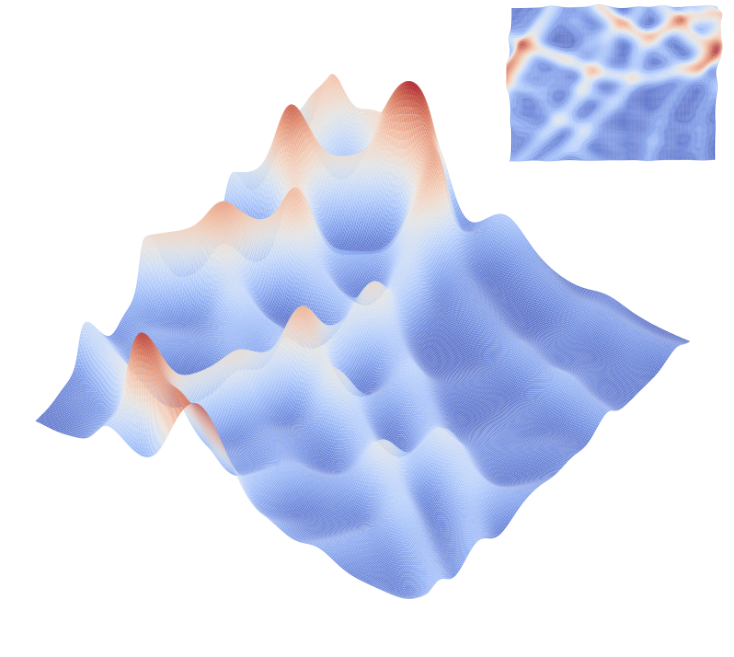}
        \caption{}
        \label{fig:terrain:seg}
    \end{subfigure}
     \begin{subfigure}[b]{0.4\textwidth}
        \includegraphics[width=\textwidth]{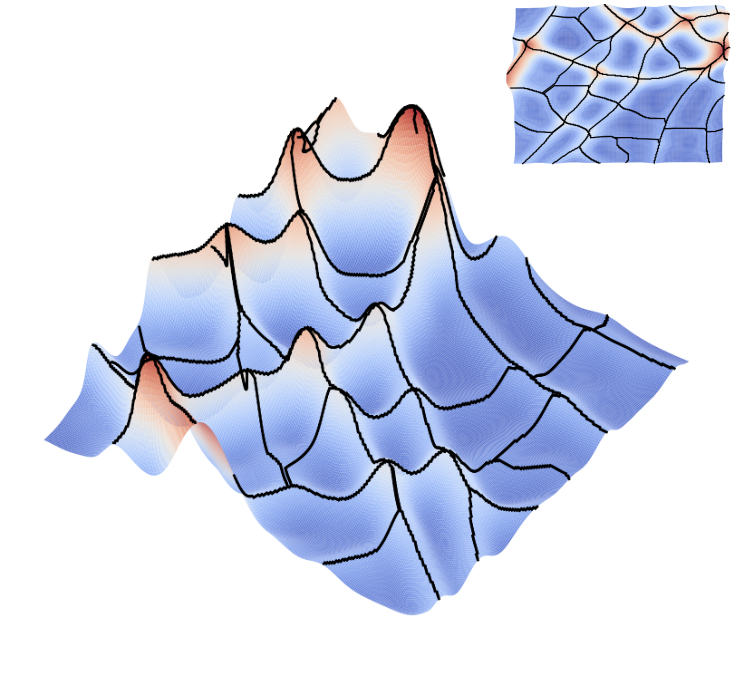}
        \caption{}
        \label{fig:terrain:ridges}
    \end{subfigure}
    \caption{(a) An input density field on the plane (top corner) and its terrain view. (b) Mountain ridges of the terrain (black lines) capture the road network. 
    \label{fig:terrain}} 
\end{figure}

Given a segmented image, we view it as a density field defined on a 2D grid, where the function value at each vertex reflects 
the likelihood of the corresponding pixel belonging to the road class. 
The goal is to extract a graph that represents the hidden road network. 

In particular, for simplicity, assume we have a triangulation $K$ of the grid (image), and thus the segmented image can be viewed as a ``density function'' $f: K \to \mathbb{R}$ with $f(s)$ reflecting how likely $s$ is in the road class.   
The graph of this density function $f$ can be viewed as a terrain with the height of a point $s$ being its function value $f(s)$; 
see Figure \ref{fig:terrain:seg}. 
Algorithm \WWLAlgSimp() of \cite{dey2018graph} (developed based on earlier work, e.g, \cite{DRS15,GDN07,2011MNRAS,wang2015efficient}) proposes to use the ``mountain ridge" of this terrain to describe the hidden graph. 
Intuitively, the mountain ridge structures are formed by those flow lines (following the steepest descending direction) that connect maxima and saddles of this terrain. Curves in the mountain ridges connect mountain peaks and saddles, and separate different ``valleys". A point on such a curve has a higher function value than points off the curve in a direction orthogonal to the curve locally. This is consistent with what a ``road" should be: points in a road have higher ``density'' than points off the road in the orthogonal direction though  this point may not have the highest density value along the road itself. See Figure \ref{fig:terrain:ridges}. 

Algorithm \WWLAlgSimp() extracts the ``mountain ridges" from the input density function (terrain) via the so-called \emph{1-stable manifolds} from Morse theory. For the sake of efficient and numerically stable computation, 
it uses the discrete Morse theory \cite{forman} to implement it.
Very importantly, the algorithm also uses the concept of \emph{persistent homology} \cite{ELZ02} to capture ``importance'' of different pieces of 1-stable manifolds (more precisely, important max-saddle pairs) in a meaningful manner. 
This allows the algorithm to remove noise and simplify the output graph (road network) systematically. 

Notice that, since this algorithm uses the global ``mountain ridges'' to infer the hidden networks, it does not need to identify the junction nodes separately, and it can also bridge through small gaps in the density field. The algorithm is clean (uses only one parameter) and efficient. It takes $O(n\log n)$ time for a planar triangulation with $n$ vertices.

%

\section{Approaches}
\label{sec:aprch}
\subsection{Semi-automatic framework}
\label{sec:semi_auto} 
\begin{figure}[t]
\includegraphics[width=\textwidth]{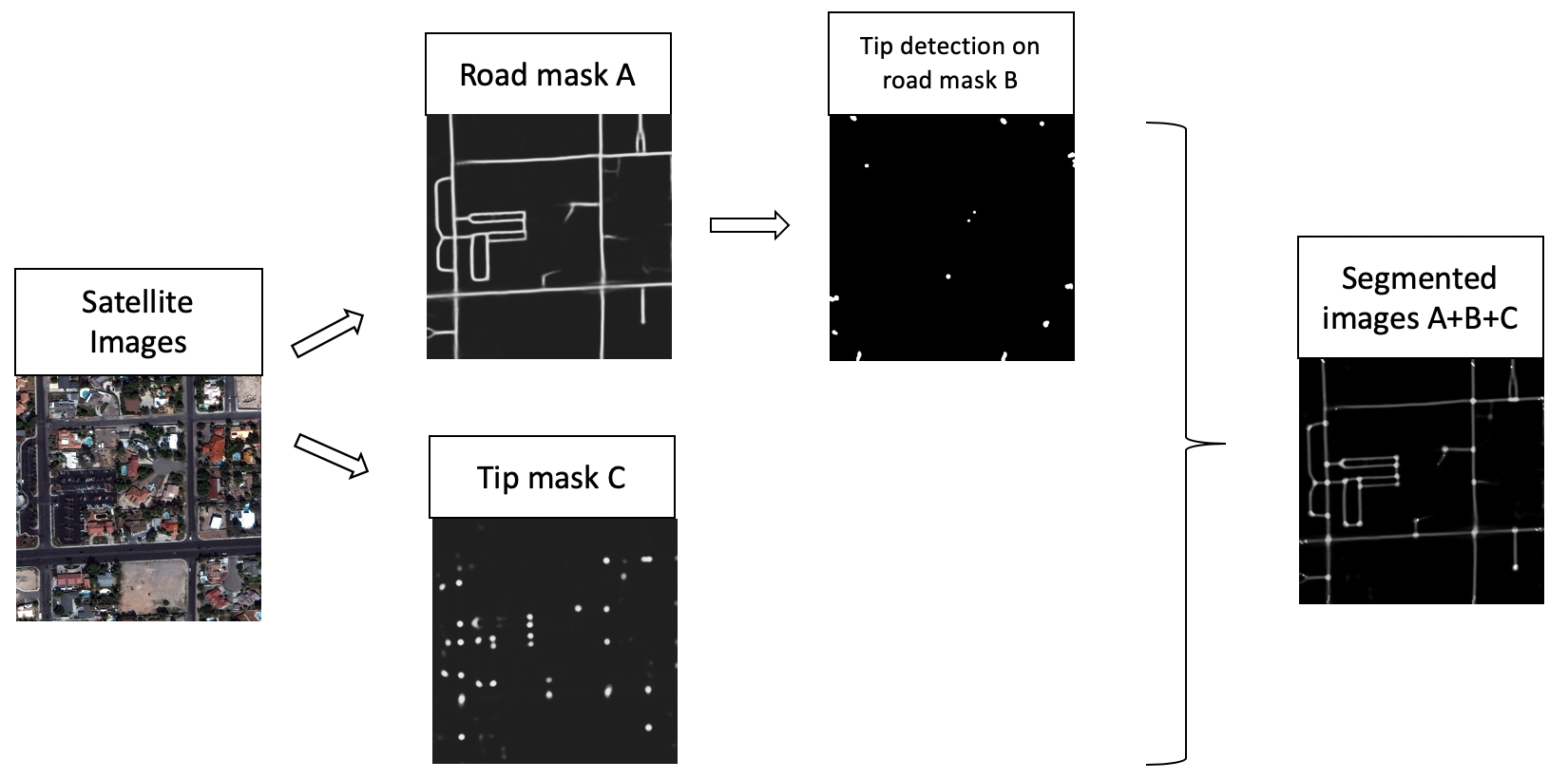}
\caption{The pipeline for Stage 1 (CNN training) for our semi-automatic framework.
\label{fig:semiautomatic}}
\end{figure}

The semi-automatic framework follows the high-level two-stage approach as outlined in Figure \ref{fig:pipeline}. 
In the first stage, we train a CNN using training images consisting of ground-truth roads labeled. 
Given a raw satellite image, we feed it to this trained CNN to obtain a segmented image. 
In the second stage, we apply the discrete-Morse based graph reconstruction algorithm to extract the road-network from the segmented image. 
For the second stage to work more accurately, we need to detect road ends called ``tips" in the segmented images obtained in the first stage.
We take advantage of the CNN to add a simple ``tip-detection'' stage that enhances the segmented images. 
The overall pipeline for Stage-1 of the semi-automatic framework is shown in Figure \ref{fig:semiautomatic}. 
%
The inputs for the framework are high resolution satellite images, which are split into a test set and a train set.
The train set has ground truth graphs (obtained manually) that represent the centerlines of the road networks.
The road-labels for training are created by thickening the ground truth graph and labeling pixels inside the thickened graph as positive and others as negative. 

\paragraph{CNN architecture}
We use the architecture from the winner's approach of the Spacenet Challenge \cite{buslaev2018fully}.
It uses  resnet34 \cite{he2016deep} as encoder and unet-like \cite{ronneberger2015u} decoder. 

\begin{figure}[H]
    \centering
    \begin{subfigure}[b]{0.3\textwidth}
        \includegraphics[width=\textwidth]{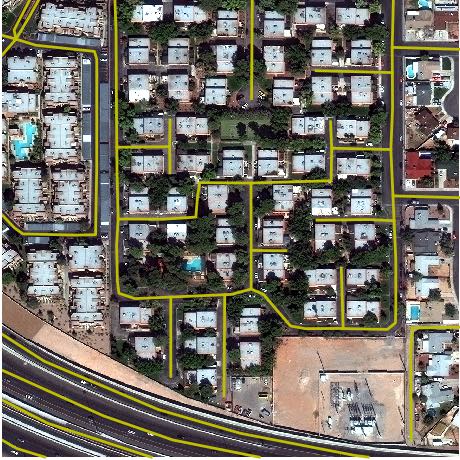}
        \caption{}
        \label{}
    \end{subfigure}
    ~ 
    \begin{subfigure}[b]{0.3\textwidth}
        \includegraphics[width=\textwidth]{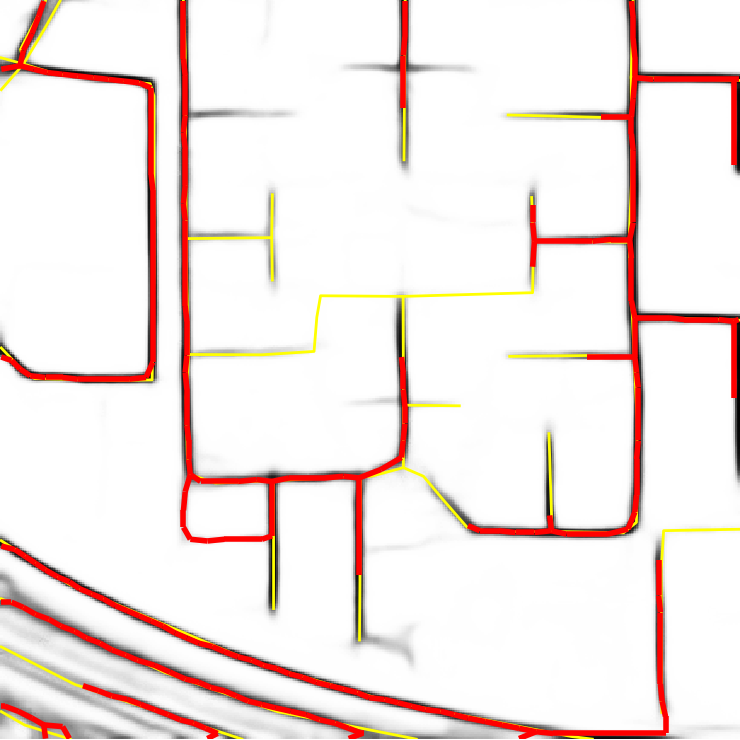}
        \caption{}
        \label{}
    \end{subfigure}
    ~ 
            \begin{subfigure}[b]{0.3\textwidth}
        \includegraphics[width=\textwidth]{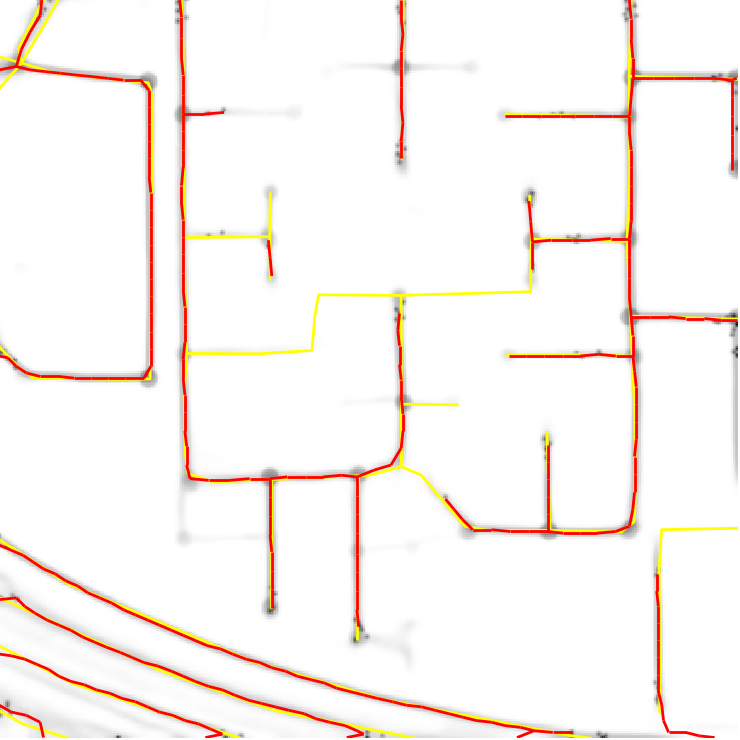}
        \caption{}
        \label{}
    \end{subfigure}
    \caption{Comparison of results without and with tip enhancements. Left: raw satellite images (yellow graphs are ground-truth road networks). Middle/Right: red-graphs are reconstructions without/with tip enhancement respectively, overlaid on top of the ground-truth graph (yellow). Dark colors are the learned density field. }\label{fig:tip-cmp}
\end{figure}

\paragraph{Reconstructing tips.}
The graph reconstruction algorithm \WWLAlgSimp() sometimes may miss hanging branches. 
To remedy this, we propose a novel way to enhance the segmented images. In particular, following the edit strategy of \cite{dey2017improved}, we modify the density values (i.e, the pixel values of the segmented images) of the tips to high values thus causing them to become local maxima which in turn forces reconstructed roads  connecting to them.
We develop two techniques to detect the tips: 
(1) Learn the locations of the tips with the same CNN architectures.
(2) Detect the tips from the segmented images by checking the windows around points with high densities.
As shown in Figure \ref{fig:semiautomatic}, we add up the segmented image and the two tip enhancements to obtain the final segmented image to feed to Stage 2. 
Figure \ref{fig:tip-cmp} shows the comparison between reconstructions without and with tip enhancements.

\subsection{Fully Automatic Framework}
\label{sec:ful_auto}

The ground truth labeling used in the semi-automatic framework is itself a graph like structure. 
In this section, we propose to create the labels using the discrete-Morse based graph reconstruction algorithm  without the knowledge of the ground truth.
These labels are used to train
a CNN for image segmentation. The segmented images are again labeled by the output of the graph reconstruction algorithm and fed to the CNN for training purpose. A few iterations of training and labeling improves the quality of the image segmentation significantly as our experiments show.
This framework is particularly useful when there is no or very few labelled data to begin with. 

\begin{figure}[t]
\includegraphics[width=\textwidth]{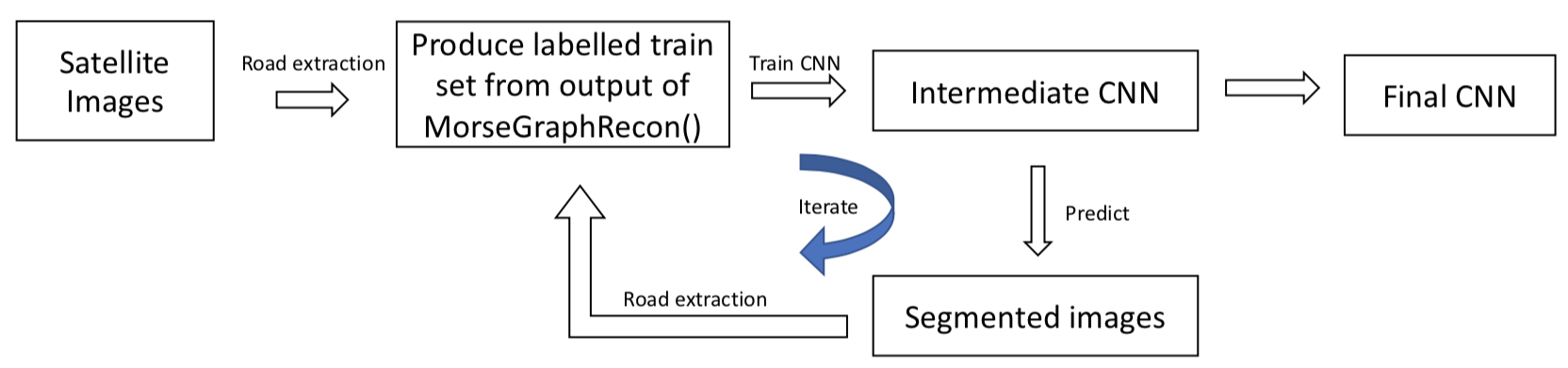}
\caption{Pipeline for Stage 1 (CNN training) for our fully automatic framework. Note that no input satellite image has labels for roads! 
\label{fig:pipeline_auto}}
\end{figure}


Our framework can deal with the following two scenarios. 
We perform {\bf label-free learning} when we do not have ground truth roads for any input satellite images to begin with.
We perform {\bf partially-labeled learning} when we have a small fraction of images (say $10\%$ of training set) with road labels.

\paragraph{Label-free case.}
We describe the framework for the label-free case, and the partially-labeled case can be handled by a slight modification of it.
The high-level pipeline of Stage 1 (training a CNN for segmenting an input image) is in Figure \ref{fig:pipeline_auto}. 
Given an input set of raw satellite images (with no labels), we split it into the training and testing sets, denoted by $I_0^{tr}$ and $I_0^{test}$, respectively. 
We run algorithm \WWLAlgSimp() on each image from $I_0^{tr}$, and let $\Ghat$ be its corresponding output. We use a large threshold for simplification in algorithm \WWLAlgSimp() so as to generate a reconstruction of the more reliable part of the input. 
Then we label pixels on $\Ghat$ as positive and pixels on the complement of $\Ghat$ as negative. 
Next we train the CNN classifier with those labeled pixels, and this is our first classifier $C_0 = \MorseTrain(I_0^{tr})$ (shown in Algorithm \ref{alg:morselabel}). 


\RestyleAlgo{boxruled}
\LinesNumbered
\begin{algorithm}[hptb]
\caption{\MorseTrain($I$) \label{alg:morselabel}}
\DontPrintSemicolon
\KwData{Images $I$}
\KwResult{Classifier $C$}
\Begin{
Compute the triangulation $K$ of $I$ and take pixel values as density function $\rho$\;

$\Ghat$ = \WWLAlgSimp($K$,$\rho$, $\delta$)\;

Label pixels on $\Ghat$ as positive, pixels on the complement of $\Ghat$ as negative\;

Train a CNN classifier $C$ by above features\;

\textbf{return} $C$\;
}
\end{algorithm}

Now feeding each original training image from $I_0^{tr}$ to $C_0$ returns a collection of segmented images $I_1^{tr}$, where in each image, every pixel has a value reflecting the likelihood of it being positive (on the road). 
We repeat the steps with images in  $I_1^{tr}$ and obtain a new CNN classifier $C_1 = \MorseTrain(I_1^{tr})$. 
In a generic $i$-th iteration of this process, feeding the training images $I_i^{tr}$ to $C_i$ returns segmented images $I_{i+1}^{tr}$, which we use to train a new CNN classifier $C_{i+1} = \MorseTrain(I_{i+1}^{tr})$.
The process terminates when the segmented images $I_i^{tr}$ undergo little changes over iterations.


\paragraph{Partially-labelled case.} 
For this scenario, we start training the CNN classifier using only 
the labelled training images to obtain $C_0$. 
In each of the subsequent iteration $i > 0$, we use both the labels computed from the segmented images $I_i^{tr}$ at this iteration, as well as the original labels from the ground truth. 

\section{Experiments}
\label{sec:exp}
\paragraph{Datasets}
We consider data from the SpaceNet Challenge 3 \cite{van2018spacenet}. 
It includes four cities: Las Vegas, Paris, Shanghai and Khartoum and consists of the original panchromatic band, the 1.24m resolution 8-band multi-spectral 11-bit geotiff, and a 30 cm resolution Pan-Sharpened 3-band and 8-band 16-bit geotiff. 
We only use the 30 cm resolution Pan-Sharpened 3-band (RGB) 16-bit geotiff in our experiments.
Each image from the dataset covers 400m by 400m with a size of 1300px by 1300px.
The ground truth for each image is a graph representing the centerline of the roads.
The width of the roads in the masks is 4 meters.
To evaluate the results, we need to compare the proposed graphs with the ground truth.
So we only take the train set from this challenge (since ground truth is only known for this set).

\paragraph{Metrics.}
The first metric we use to evaluate the results is the Average Path Length Similarity (APLS) \cite{van2018spacenet}. This is the metric used for evaluation in SpaceNet Challenge 3. 

\begin{definition}
Let $G_1=(V_1,E_1)$ and $G_2=(V_2,E_2)$ be two input graphs.
For $a,b\in V_1$ where $path(a,b)$ exists in $G_1$, 
let $a'$ (resp. $b'$) denote the closet node to $a$ (resp. to $b$) in $G_2$. 
$L(\cdot,\cdot)$ denote the length of the shortest path.
First we define the cost of $path(a,b)$:
\[ c(a,b)=
\begin{cases} 
      min\left\{  1,\frac{|L(a,b)-L(a',b')|)}{L(a,b)}\right\}, & if \:path(a',b')\: exists\\
      1, & otherwise \\
   \end{cases}
\]
We next define
\begin{equation*}
C(G_1,G_2) = 1- \frac{1}{N}\sum c(a,b)
\end{equation*}
Where $N$= \# unique paths in $G_1$, 
and we take the sum over all unique paths.
Finally, the \emph{APLS score of $G_1$ and $G_2$} is defined to be the  harmonic mean of $C(G_1,G_2)$ and $C(G_2,G_1)$:
\begin{equation*}
APLS(G_1,G_2)= \frac{2}{\frac{1}{C(G_1,G_2)}+\frac{1}{C(G_2,G_1)}}
\end{equation*}
\end{definition}

\begin{figure}[htbp]
    \centering
        \begin{subfigure}[b]{0.3\textwidth}
        \includegraphics[width=\textwidth]{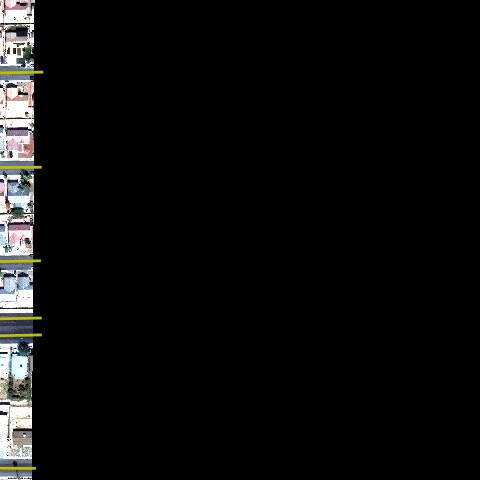}
        \caption*{AOI\_2-Id: 1634}
        \label{}
    \end{subfigure}
    \begin{subfigure}[b]{0.3\textwidth}
        \includegraphics[width=\textwidth]{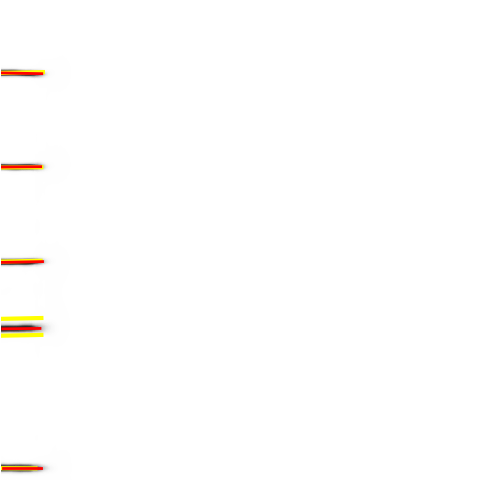}
        \caption*{0.8560 / 14.0662}
        \label{}
    \end{subfigure}
    \begin{subfigure}[b]{0.3\textwidth}
        \includegraphics[width=\textwidth]{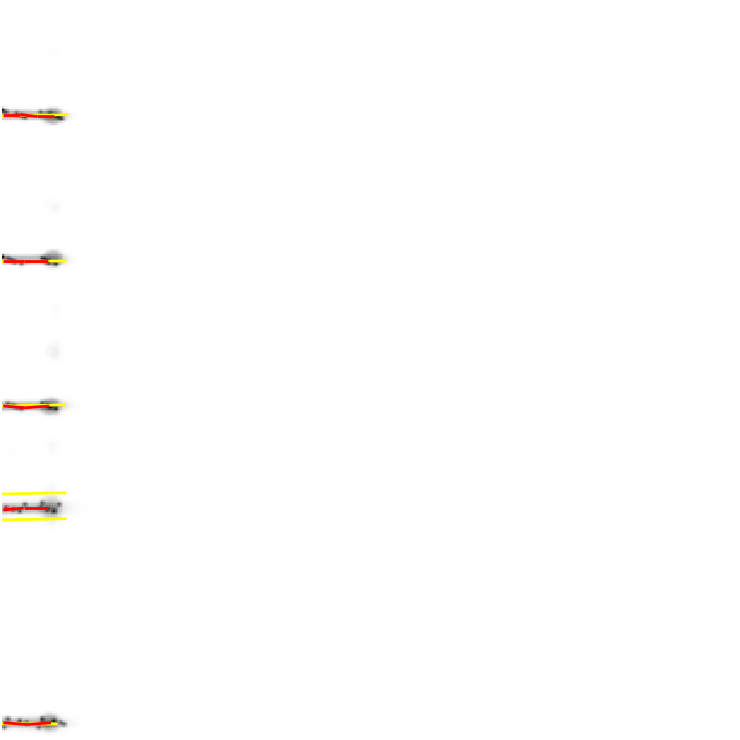}
        \caption*{0.4684 / 15.3238}
        \label{}
    \end{subfigure}
    \caption{\label{fig:mt}Left: input satellite image. Middle/Right: reconstruction of Buslaev's method and our method. The APLS-score/\Hdis are listed. Note that even though our reconstruction is very similar, the APLS-score is significantly lower due to the sparsity of the signal. \Hdis is more accurate for this case.}
\end{figure}

This metric sums the differences in optimal paths existing between nodes in the ground truth graph $G_1$ and the reconstructed graph $G_2$.
It consists of two parts:
part 1 considers optimal paths from the ground truth graph, finds paths from the reconstructed graph which correspond to them and measure differences; 
and part 2, in opposite direction, considers paths from the reconstructed graph, finds their correspondences in the ground-truth and compare them.
The final score is the harmonic mean of these two parts.
It cares about the connections between the nodes and punishes breaks in the roads. 
However, this metric may not be accurate when the size of the graph and the total amount of paths are small since the metric evaluates the portion (ratio) of paths that match well.
In this case, a small difference in the graphs could result in a relatively large difference in the score (see Figure \ref{fig:mt}).
To obtain a more comprehensive picture, we also use the following \Hdis: 
\begin{definition}\label{def:hau-normal}
Suppose $G_1$ and $G_2$ are two graphs; $P_1$ is the point set sampled from $G_1$; $P_2$ is the point set sampled from $G_2$, 
and $d$ denotes the Euclidean distance. Then, the one-directional Hausdorff distance is: 
\[ S_{HA}(G_1,G_2):=
\begin{cases} 
      \frac{1}{|P_1|}\sum_{p\in P_1}d(p,P_2), &G_1\neq\emptyset \;and\; G_2\neq\emptyset\\
      MAX, & Only \;one\; graph\; is\; empty\\
      0, & G_1=G_2=\emptyset
   \end{cases}
\]
Here, MAX is a specific maximum value.  
We set  $S_H(G_1, G_2) = S_{HA}(G_1,G_2)+ S_{HA}(G_2,G_1)$ as final \emph{\Hdis} between $G_1$ and $G_2$.
\end{definition}


Note that for APLS score, the higher the score is, the more similar the two graphs are. 
But for \Hdis, the lower the distance is, the more similar the two graphs are. 

\paragraph{Parameters. }
There are several parameters in the entire pipeline, among which the persistence threshold $\delta$ (for the discrete-Morse based reconstruction algorithm) and the arc-intensity threshold $\tau$ (used to further remove noisy arcs during the post-processing) affect the results most.
To tune these two hyperparameters, we experiment on the validation set with a range of parameters that are chosen empirically, then take the set of parameters that give the highest score.
We take APLS scores as the reference to tune the parameters. 

Furthermore, for datasets AOI\_3 (Shanghai) and AOI\_4 (Paris), there are many images with extremely sparse signal, while many of them have much denser signal. We thus use a two-threshold system for the arc-intensity threshold:  
For those images we need a low arc-intensity threshold $\tau$: We sort the images by the sum of their intensities, and apply a lower $\tau$ to those images with low total-intensity. We use a higher $\tau$ for the remaining images. 
\begin{wrapfigure}{r}{3cm}
        \includegraphics[height=2.5cm]{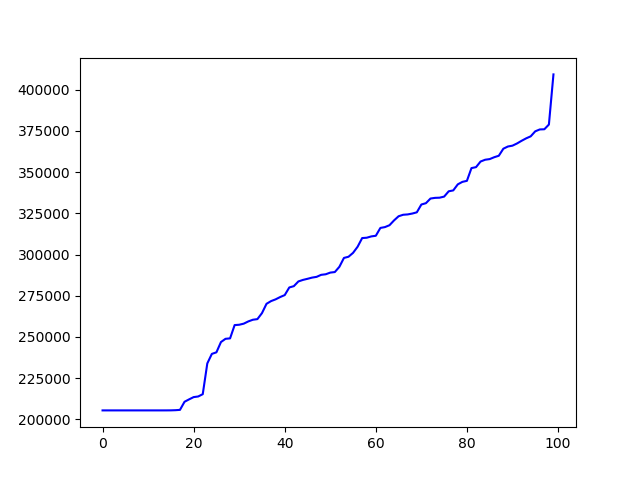}
\end{wrapfigure}
For example, see the right figure for dataset AOI\_4, where $x$-axis is the percentage of images (sorted in increasing total-intensity), and $y$-axis is their total-intensity. Given that there is a sharp transition around $20\%$, we apply a lower threshold to the $20\%$ of images with the lowest total intensity. We use the same strategy for AOI\_3, and choose $40\%$ as the threshold to have two $\tau$ values. 

%

\subsection{Semi-automatic reconstruction results}
\label{sec:semi-exp}
\noindent

\begin{table}[htbp]
\begin{center}
  \begin{tabular}{| l | l | c | l | l |}
    \hline
     & train & validation& test & total\\ \hline
    AOI\_2\_Vegas & 659 &165 &165& 989\\ \hline
    AOI\_3\_Paris &206  & 52 &52&310\\ \hline
    AOI\_4\_Shanghai & 798 & 200 &200&1198\\ \hline
    AOI\_5\_Khartoum &189 & 47 &47&283\\ \hline
          \end{tabular}
          \caption{The split of the dataset.The ratio for train/test/valid data is approximately 4:1:1.}\label{tab:split}
\end{center}
\end{table}

\begin{table}[htbp]
\centering
\begin{tabular}{|l|l|l|l|l|}
\hline
     & \multicolumn{2}{l|}{APLS}                                           & \multicolumn{2}{l|}{$S_H$} \\ \hline
     & Buslaev\cite{albu}                           & ours                             & Buslaev\cite{albu}        & ours       \\ \hline
AOI\_2 & 0.8211 & \textbf{0.8278}                          &18.3539               &\textbf{17.7841}            \\ \hline
AOI\_3 & 0.5848 & \textbf{0.6324}                          &291.0188               & \textbf{289.9532}           \\ \hline
AOI\_4 & 0.6630                           & \textbf{0.6632} &69.5775               &\textbf{68.9596}            \\ \hline
AOI\_5 & 0.6069                           & \textbf{0.6477} & 44.4201              &\textbf{41.6037}            \\ \hline
\end{tabular}
\caption{APLS score and \Hdis for the test set. The $MAX$ value used for \Hdis is 500 pixels, the size of each image is 1300px by 1300px. The \Hdis for AOI\_3 is high because there are more cases where the proposed graph is empty while the ground truth graph is not.}
\label{tab:two_scores}
\end{table}

\begin{table}[htbp]
\centering
\begin{tabular}{|l|l|l|}
\hline
       & $\delta$ & $\tau$ \\ \hline
AOI\_2 &0.12          & 0.4       \\ \hline
AOI\_3 & 0.1         &0.3(30\%)/0.4        \\ \hline
AOI\_4 &  0.1        & 0.3(40\%)/0.4       \\ \hline
AOI\_5 &  0.07        & 0.3       \\ \hline
\end{tabular}
\caption{Chosen parameters. For AOI\_3, the content in the column $\tau$ means that for 40\% of images with the lowest total intensity, take $\tau=0.3$ and for the rest of the images take $\tau=0.4$. The same for AOI\_4.}
\label{tab:para}
\end{table}

\paragraph{Compared method: Buslaev's method \cite{albu}}
We compare our framework with the method of the winner of SpaceNet Challenge 3 \cite{albu}. 
It uses the same CNN architecture to train and then predict the segmented images. 
In Stage 2, Buslaev's method first extracts the skeleton from the thresholded segmented images. 
Then, it transforms the skeleton to a multi-graph using library ``sknw'' \cite{sknw}. Finally, it translates the multi-graph to a graph with straight edges. 
Buslaev's method outperforms other methods in SpaceNet Challenge 3, so we only compare ours with this one.

\paragraph{Results.} 
As mentioned before, we tested on four datasets. The split of train-validation-test is 4:1:1 for each data set, and the precise  numbers are listed in Table \ref{tab:split}. 

\begin{figure}[htbp]
    \centering
 \begin{subfigure}[b]{0.3\textwidth}
        \includegraphics[width=\textwidth]{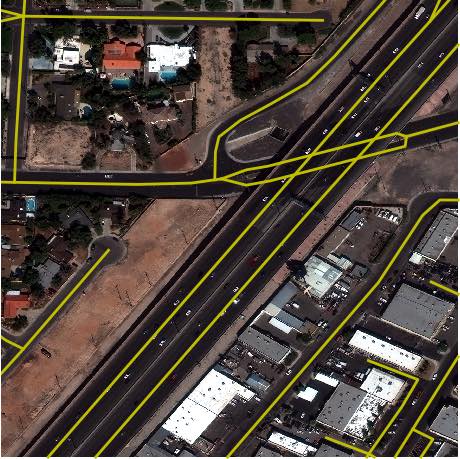}
        \caption*{AOI\_2-Id: 1462}
        \label{}
    \end{subfigure}
    ~ 
    \begin{subfigure}[b]{0.3\textwidth}
        \includegraphics[width=\textwidth]{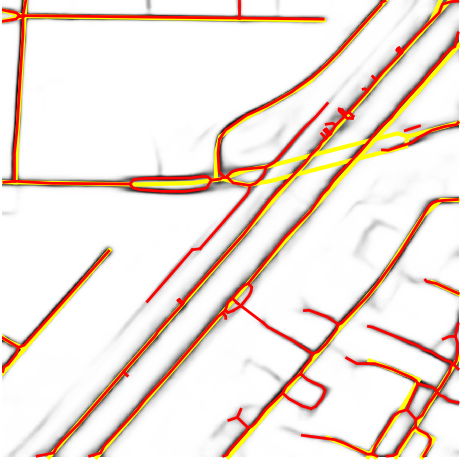}
        \caption*{0.5700 / 18.6828 }
        \label{}
    \end{subfigure}
    ~
        \begin{subfigure}[b]{0.3\textwidth}
        \includegraphics[width=\textwidth]{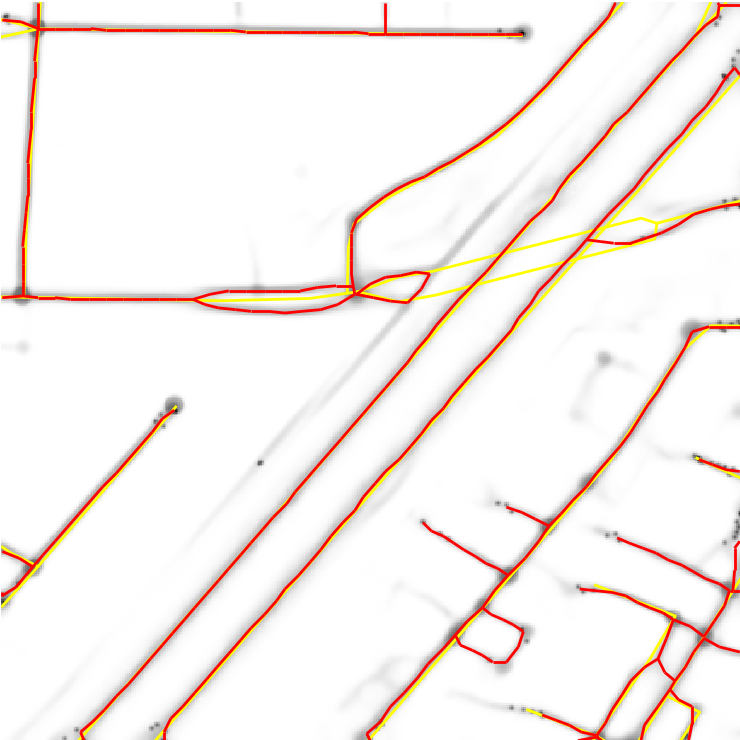}
        \caption*{0.6655 / 14.2246  }
        \label{}
    \end{subfigure} \\
    \begin{subfigure}[b]{0.3\textwidth}
        \includegraphics[width=\textwidth]{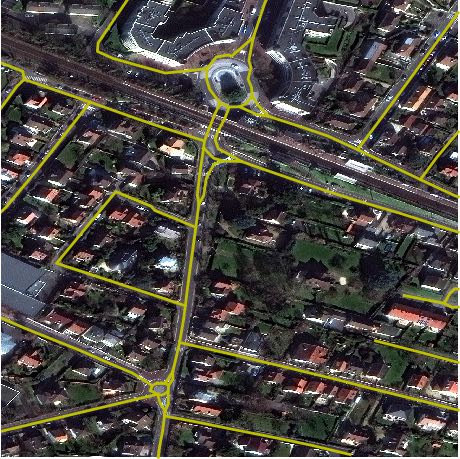}
        \caption*{AOI\_3-Id: 217}
        \label{}
    \end{subfigure}
    ~ 
    \begin{subfigure}[b]{0.3\textwidth}
        \includegraphics[width=\textwidth]{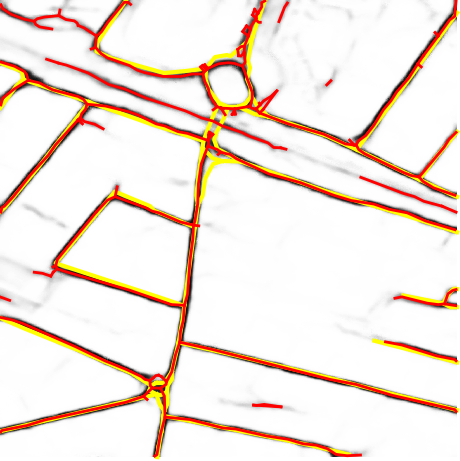}
        \caption*{0.6194 / 20.7057 }
        \label{}
    \end{subfigure}
    ~
        \begin{subfigure}[b]{0.3\textwidth}
        \includegraphics[width=\textwidth]{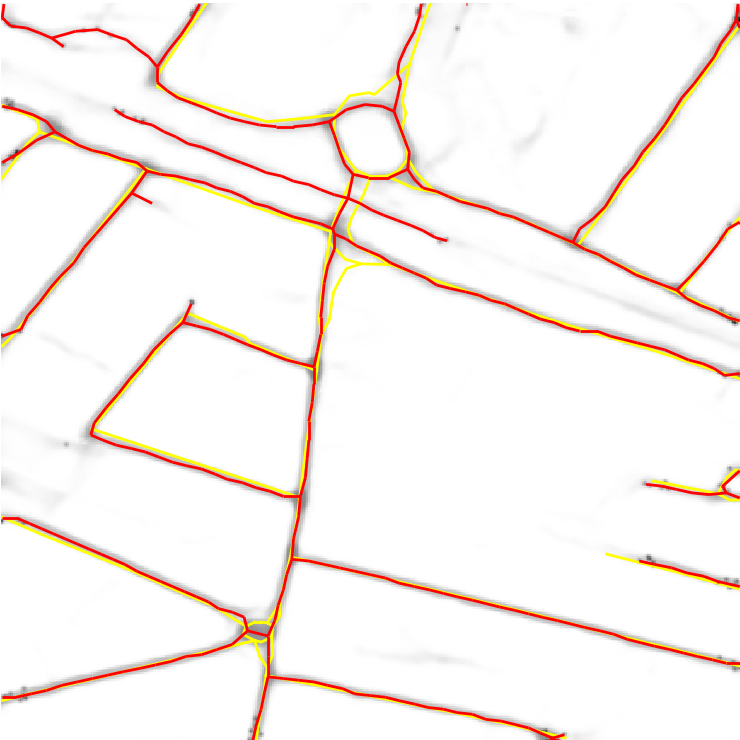}
        \caption*{0.8193 / 16.8086  }
        \label{}
    \end{subfigure}
    \\
    \begin{subfigure}[b]{0.3\textwidth}
        \includegraphics[width=\textwidth]{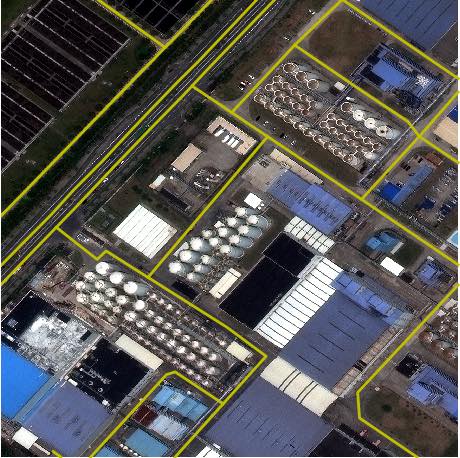}
        \caption*{AOI\_4-Id: 267}
        \label{}
    \end{subfigure}
    ~ 
    \begin{subfigure}[b]{0.3\textwidth}
        \includegraphics[width=\textwidth]{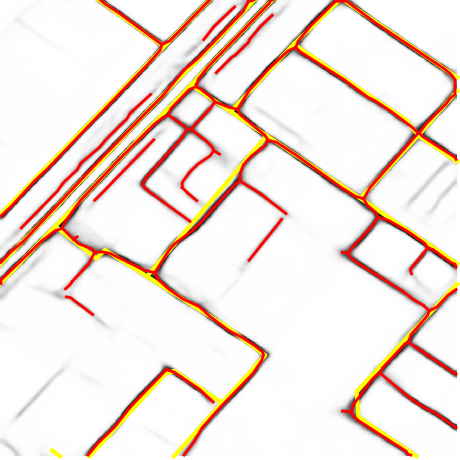}
        \caption*{0.4721 / 29.5151 }
        \label{}
    \end{subfigure}
    ~
        \begin{subfigure}[b]{0.3\textwidth}
        \includegraphics[width=\textwidth]{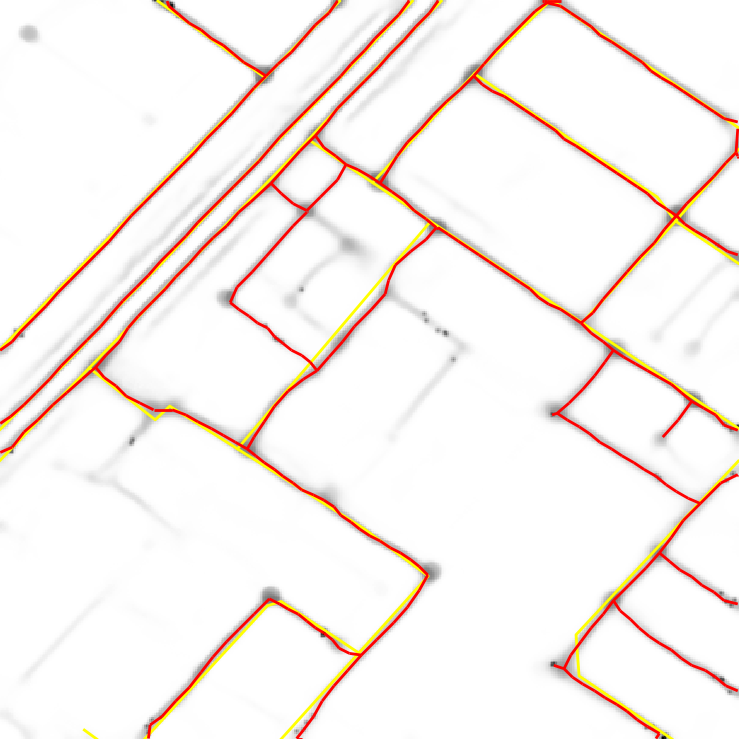}
        \caption*{0.5693 / 21.6930  }
        \label{}
    \end{subfigure}
    \\
       \begin{subfigure}[b]{0.3\textwidth}
        \includegraphics[width=\textwidth]{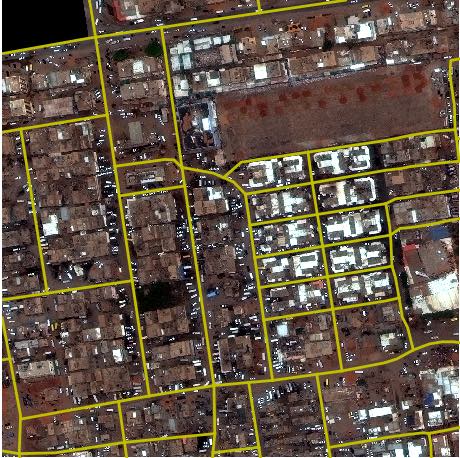}
        \caption*{AOI\_5-Id: 207}
        \label{}
    \end{subfigure}
    ~ 
    \begin{subfigure}[b]{0.3\textwidth}
        \includegraphics[width=\textwidth]{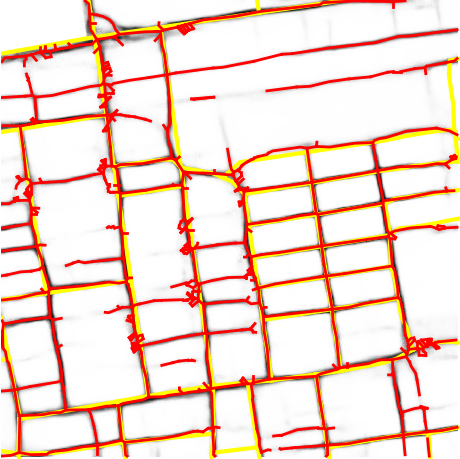}
        \caption*{0.6334 / 30.0484 }
        \label{}
    \end{subfigure}
    ~
        \begin{subfigure}[b]{0.3\textwidth}
        \includegraphics[width=\textwidth]{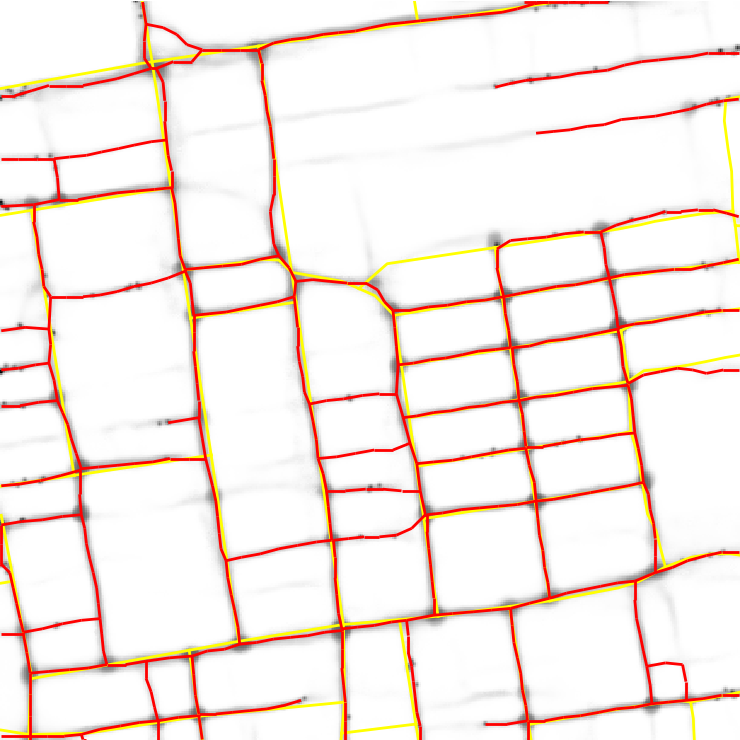}
        \caption*{0.7287 / 24.9911  }
        \label{}
    \end{subfigure}
    \caption{Left: raw satellite images (yellow graphs are ground-truth road networks). Middle/right: red-graphs are reconstruction of Buslaev's method and our semi-automatic framework, respectively, overlaid on top of the ground-truth graph (yellow). Dark colors are the learned density field. The numbers given are APLS-score/\Hdis.  \label{fig:reconexamples}}
\end{figure}

Tables \ref{tab:two_scores} shows the scores under the two metrics for Buslaev's framework and ours over test datasets (on validation datasets our scores are consistently better). 
Each score is an average of scores for all test images (recall the split of train-validation-test is shown in Table \ref{tab:split}). 
For APLS score, the larger value the better it is. 
For \Hdis, the smaller value the better it is.
Note that AOI\_4 and AOI\_5 are rather noisy images (especially AOI\_5) and most challenging among all datasets. 
Our method significantly outperforms Buslaev's method on AOI\_5. 
We also observe that, in general, our output tends to have better connectivity. 
Figure \ref{fig:reconexamples} shows a few examples. 
Buslaev's algorithm tends to have more extra branches, and worse connectivity. 
We note that the final average APLS score reported here for Buslaev's method is different from the posted one 0.6663 in \cite{spacenetResult}. This is because the posted score is computed for the original test set from SpaceNet challenge, while we our test set is a subset of the original train set -- we cannot compare on the original test set from SpaceNet challenge as the ground truth for them are not publicly available.
Tables \ref{tab:para} shows the finally chosen parameters for the reproducibility of the experiment.

\paragraph{Running time.}
For each of the two larger datasets (AOI\_2\_Vegas and AOI\_3\_Shanghai): Training takes around 500 minutes to learn both the lanes and the tip-marks. Testing (to obtain the segmentation on all images from the test-set) takes around 18 minutes. The final road network extraction stage takes 20 mins for each choice of two hyper-parameters (persistence threshold and intensity threshold), and total 20 * 9 = 180 minutes to tune the two hyper parameters on validation sets and then run on the final testing sets. We note that the graph reconstruction code is not optimized and we believe can be improved for the 2D setting, which would improve the time for the last final road network extraction stage. 

For each of the two smaller datasets (AOI\_3\_Paris and AOI\_5\_Khartoum): Training takes around 140 minutes to learn both the lanes and tip-marks. Testing (to obtain the segmentation on all images from the test-set) takes around 6 minutes. The final road network extraction stage takes 6 mins for each choice of parameter set, and total 54 minutes to tune the two hyper parameters on validation sets and then run on the final testing sets. 

\begin{table}[htbp]
\centering
\small\addtolength{\tabcolsep}{-3pt}
\begin{tabular}{|c|c|c|c|}
\hline
       & Train & Test & Road extraction \\ \hline
AOI\_2 & $2\times 242m$   & $8m\times 2$   & $22m\times 9$             \\ \hline
AOI\_3 & $77m\times 2$   & $2m\times 2$   & $6m\times 9$              \\ \hline
AOI\_4 & $293m\times 2$  & $10m\times 2$  & $27m\times 9$             \\ \hline
AOI\_5 & $71m\times 2$   & $2m\times 2$   & $6m\times 9$              \\ \hline
\end{tabular}
\caption{The running time: $m$ stands for \emph{minutes}; $2\times$ means it will be run twice (for both road and tip detections). $9\times$ comes from the tuning of the parameters.}
\label{tab:time}
\end{table}
 
\subsection{Fully automatic reconstruction results}

In the following experiments, we randomly select 200 images as the training set $I_0^{tr}$, and 50 images as the test set $I_0^{test}$ for each dataset. 
We evaluate the method by computing the APLS scores on the original test set after each iteration. 
We initialize our fully automatic approach by converting each RGB image to grayscale and then applying a Gaussian filter. 
One could potentially use other image processing methods to further pre-process it. 
When applying the graph reconstruction algorithm, we use the same parameters used in Section \ref{sec:semi-exp} Table \ref{tab:para}.

\paragraph{Alternative method for centerline detection.}
To show that the discrete-Morse based graph reconstruction algorithm is important for our fully-automatic training framework, we develop the following alternative scheme  \SkeTrain() as a baseline to compare: the graph reconstruction algorithm is replaced with the Buslaev's \cite{albu} skeleton extraction algorithm (as described in Section \ref{sec:semi-exp}).

\begin{figure}[htbp]
    \centering
        \begin{subfigure}[b]{0.3\textwidth}
        \includegraphics[width=\textwidth]{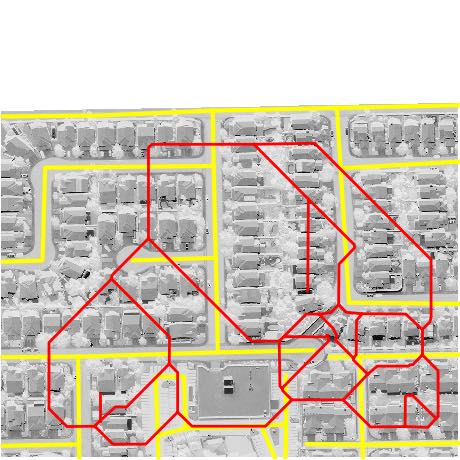}
        \caption{ }
        \label{}
    \end{subfigure}
    \begin{subfigure}[b]{0.3\textwidth}
        \includegraphics[width=\textwidth]{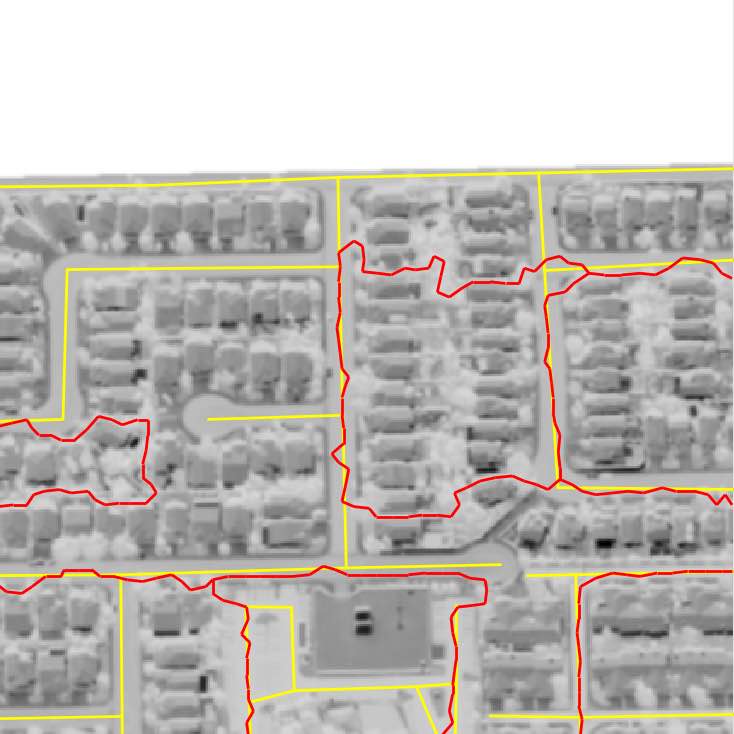}
        \caption{}
        \label{}
    \end{subfigure}
    \caption{Left: Skeleton by \cite{albu}. Right: Skeleton by discrete-Morse algorithm.}
    \label{fig:ske}
\end{figure}

Note that this skeleton extraction used in \cite{albu} is not designed to work directly on the raw satellite images; see Figure \ref{fig:ske}, where in (a) we show an output by this skeleton extraction algorithm directly applied to a raw satellite image (yellow curves are ground truth), while (b) shows the output of the discrete Morse-based algorithm on the same input,
which is much better. 
%
Hence to improve the performance of this the baseline method \SkeTrain(), we will still first use the discrete-Morse graph reconstruction algorithm (or if there are partially-labelled data, using those first) at the beginning of the training process, and switch to the skeleton extraction algorithm only after a few iterations. 

\begin{table*}[htbp]
\resizebox{\columnwidth}{!}{%
\begin{tabular}{|l|ll|l|l|l|l|lllll}
\hline
AOI\_2                         & \multicolumn{1}{l|}{$I_1^{test}$} & $ I_2^{test}$ & $ I_3^{test} $ & $ I_4^{test}$ & $I_5^{test}$ & $I_6^{test}$ & \multicolumn{1}{l|}{$I_7^{test}$} & \multicolumn{1}{l|}{$I_8^{test}$} & \multicolumn{1}{l|}{$I_9^{test}$} & \multicolumn{1}{l|}{$I_{10}^{test}$} & \multicolumn{1}{l|}{$I_{11}^{test}$} \\ \hline
\MorseTrain() & \multicolumn{1}{l|}{0.2523}     & 0.3340      & 0.3886       & 0.4173      & 0.4332                         & 0.4655                         & \multicolumn{1}{l|}{0.4829}                         & \multicolumn{1}{l|}{0.5252}                         & \multicolumn{1}{l|}{0.5497}                         & \multicolumn{1}{l|}{0.5813}                          & \multicolumn{1}{l|}{0.5922}                          \\ \hline
\SkeTrain()   &                                 &             & 0.2677       & 0.2643      & 0.2763                         & 0.2753                         &                                                     &                                                     &                                                     &                                                      &                                                      \\ \cline{1-1} \cline{4-7}
\end{tabular}
}
\caption{APLS-score for the reconstructed road networks for testing images, based on our label-free framework (\MorseTrain()), compared to the alternative method \SkeTrain(). The first two iterations for \SkeTrain() is done by \MorseTrain() and thus are not shown. After 6 iterations, the score does not improve for \SkeTrain() any more.  
\label{tab:label-free}}
\end{table*}

\begin{figure}[htbp]
    \centering
        \begin{subfigure}[b]{0.3\textwidth}
        \includegraphics[width=\textwidth]{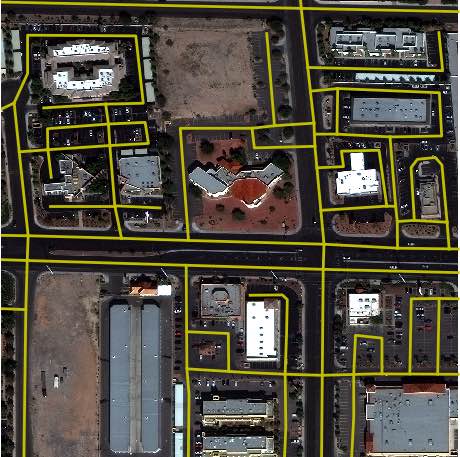}
        \caption*{AOI\_2 - Id: 323 }
        \label{}
    \end{subfigure}
    \begin{subfigure}[b]{0.3\textwidth}
        \includegraphics[width=\textwidth]{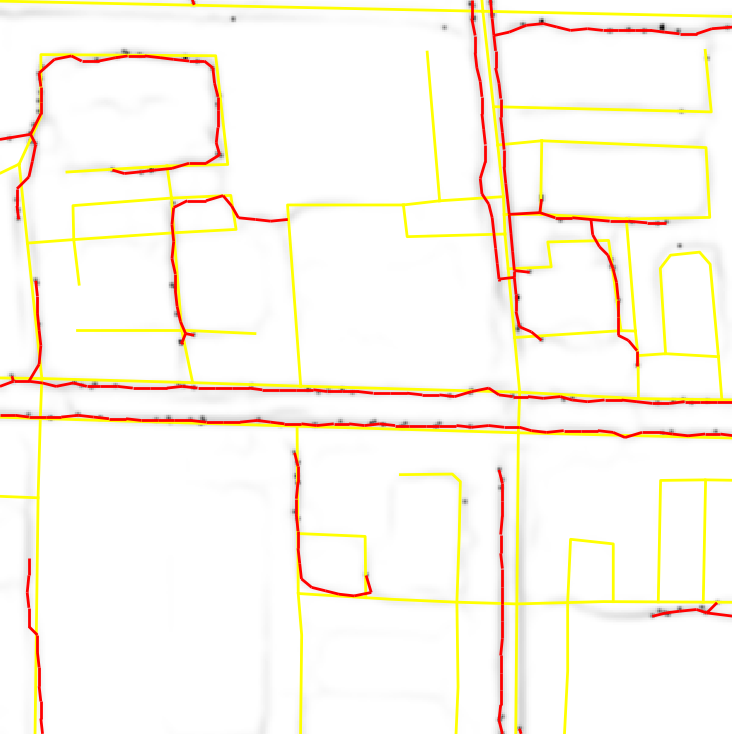}
        \caption*{$I^{te.}_1$ 0.0653 / 57.0400}
        \label{}
    \end{subfigure}
        \begin{subfigure}[b]{0.3\textwidth}
        \includegraphics[width=\textwidth]{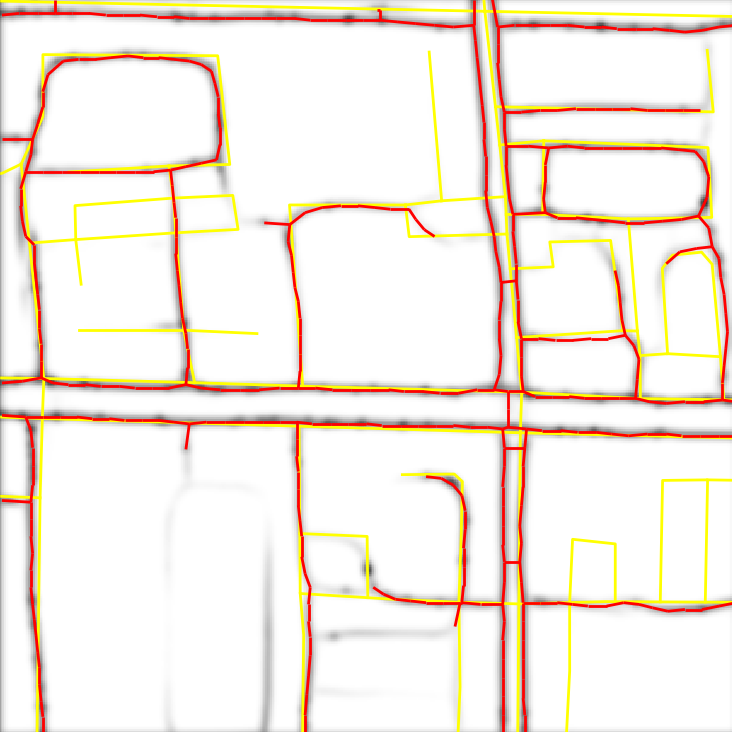}
        \caption*{$I^{te.}_{11}$ 0.5564 / 30.3855}
        \label{}
    \end{subfigure}
    \caption{Left: raw satellite image (ground truth in yellow, even though it is not used!). Middle / right: the reconstructed graph using CNN after the first and the 11-th iterations. }
    \label{fig:aoi2_n_gt}
\end{figure}

\paragraph{Results for label-free case.}
We show results here for dataset AOI\_2\_Vegas, which is a cleaner dataset from SpaceNet Challenge. Our new fully-automatic framework is less effective on AOI\_5\_Khartoum, which is much more noisy; however, we will show later that, with 10\% labelled images, it can obtain reasonable results on the challenging AOI\_5\_Khartoum dataset as well.  

For test images, we always apply tip detection and arc removal when running the graph reconstruction algorithm. 
These two procedures are not applied to the segmented images during the first three iterations of the training process of the pipeline in Figure \ref{fig:pipeline_auto}, as removing arcs results in loss of signals and tip detection tends to introduce noise when the segmented images are not yet reliable. 
From $I_4^{tr}$ onward, we start to apply tip detection since the segmented images are now less noisy. 
We also decrease the threshold for persistence simplification for the discrete-Morse based graph reconstruction for $i \ge 8$, as the quality of segmented images becomes better and better. 

In Table \ref{tab:label-free}, we show the APLS-score for test images using the CNN $C_i$ learned at the $i$-th iterations, as $i$ increases (the \Hdis shows a similar trend). 
In particular, $I_i^{test} (i>0)$ represents the output reconstructed from the segmented images of the set $I_0^{test}$ using the trained CNN $C_{i-1}$.
We compare the output of our framework for label-free case, denoted by \MorseTrain(), with the output of the baseline method \SkeTrain(). 
Note that, as explained earlier, the first two iterations for \SkeTrain() are done by \MorseTrain() (using discrete Morse graph reconstruction), and thus no APLS-scores are given for those two iterations for \SkeTrain(). 
Also no APLS-scores is shown for \SkeTrain() after the 6th iterations as the score does not improve further. 
In contrast, the APLS-score continues to improve (for test images) during the iterative process. 
In Figure \ref{fig:aoi2_n_gt}, we show an example of the reconstructed graph using the CNN from different iterations of our fully automatic training process: observe that at the beginning, only part of signals are captured. Subsequently, the classifier becomes better and more and more signals are captured. 

For this set of (200 + 50) images sampled from AOI\_2 dataset, the APLS-score for our semi-automatic framework is about $0.796$. In this fully automatic framework, in the end we obtain a score of $0.592$, which is worse. However, keep in mind that no labels are used at all. 

\begin{table}[]
\centering
\begin{tabular}{|l|l|l|l|l|}
\hline
 APLS         & $I_1^{test}$ & $I_2^{test}$ & $I_3^{test}$ & $I_4^{test}$ \\ \hline
AOI\_2 ours  & 0.6521     & 0.6918     & 0.7305     & 0.7210     \\ \hline
AOI\_2 Skel. & 0.4860     & 0.5137     & 0.5214     & 0.5252     \\ \hline\hline
AOI\_5 ours  & 0.5351     & 0.5787     & 0.5893     & 0.6077     \\ \hline
AOI\_5 Skel. & 0.5247     & 0.5091     & 0.4884     & 0.4543     \\ \hline
\end{tabular}
\caption{APLS score for partially-labeled case, where 10\% random images have road labels.}
\label{tab:10gt_APLS}
\end{table}


\begin{figure}[H]
    \centering
                \begin{subfigure}[b]{0.3\textwidth}
        \includegraphics[width=\textwidth]{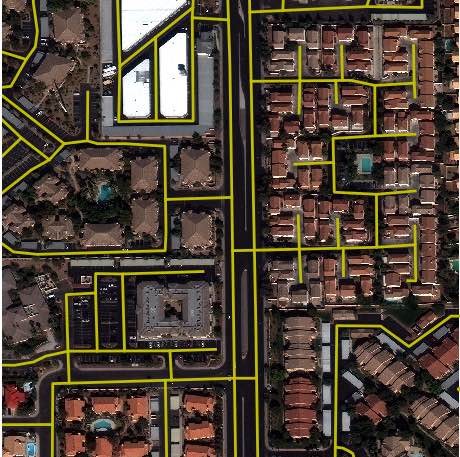}
        \caption*{AOI\_2 - Id: 429}
        \label{}
    \end{subfigure}
    \begin{subfigure}[b]{0.3\textwidth}
        \includegraphics[width=\textwidth]{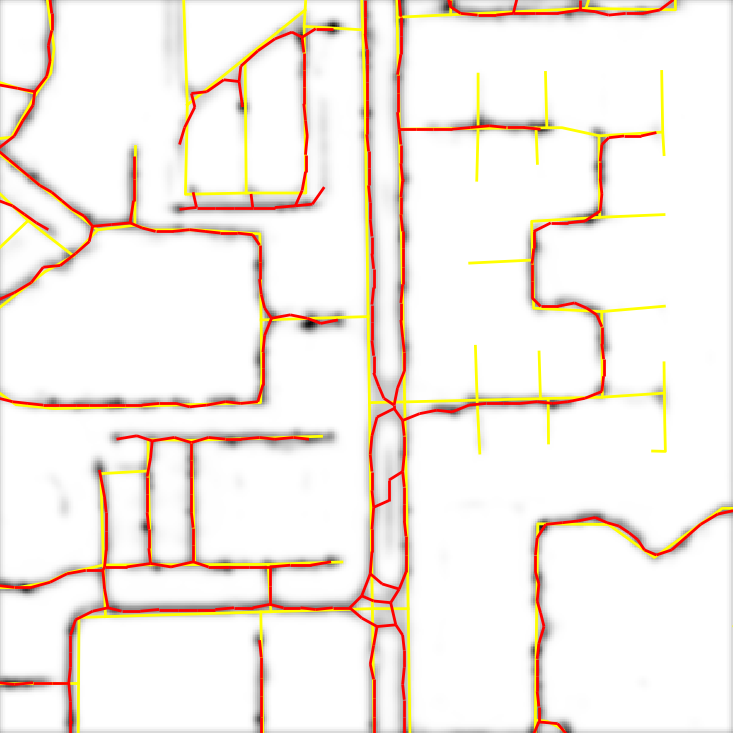}
        \caption*{$I_1^{te.}$ 0.3926 / 18.3656}
        \label{}
    \end{subfigure}
    \begin{subfigure}[b]{0.3\textwidth}
        \includegraphics[width=\textwidth]{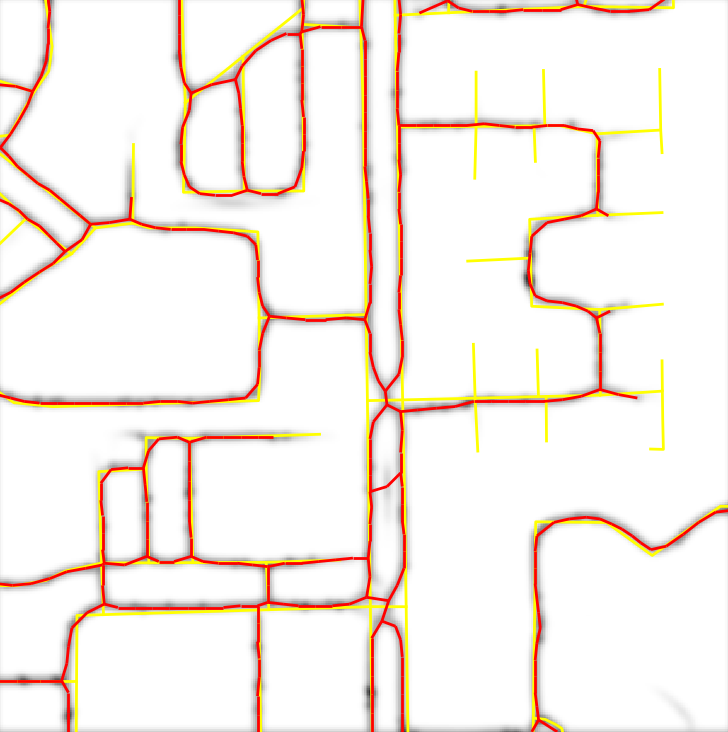}
        \caption*{$I_4^{te.}$ 0.6374 / 15.6549}
        \label{}
    \end{subfigure}
    \\
            \begin{subfigure}[b]{0.3\textwidth}
        \includegraphics[width=\textwidth]{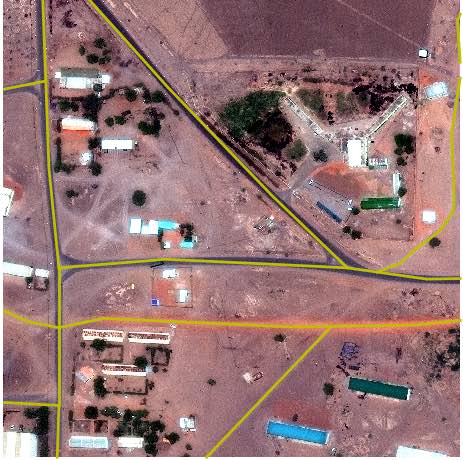}
        \caption*{AOI\_5 - Id: 50}
        \label{}
    \end{subfigure}
    \begin{subfigure}[b]{0.3\textwidth}
        \includegraphics[width=\textwidth]{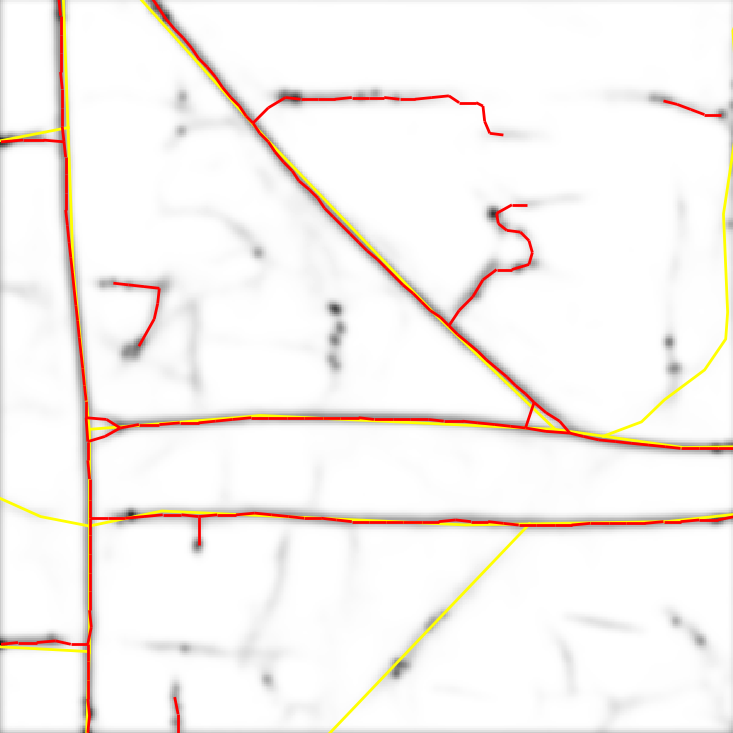}
        \caption*{$I_1^{te.}$ 0.6449 / 68.8017}
        \label{}
    \end{subfigure}
    \begin{subfigure}[b]{0.3\textwidth}
        \includegraphics[width=\textwidth]{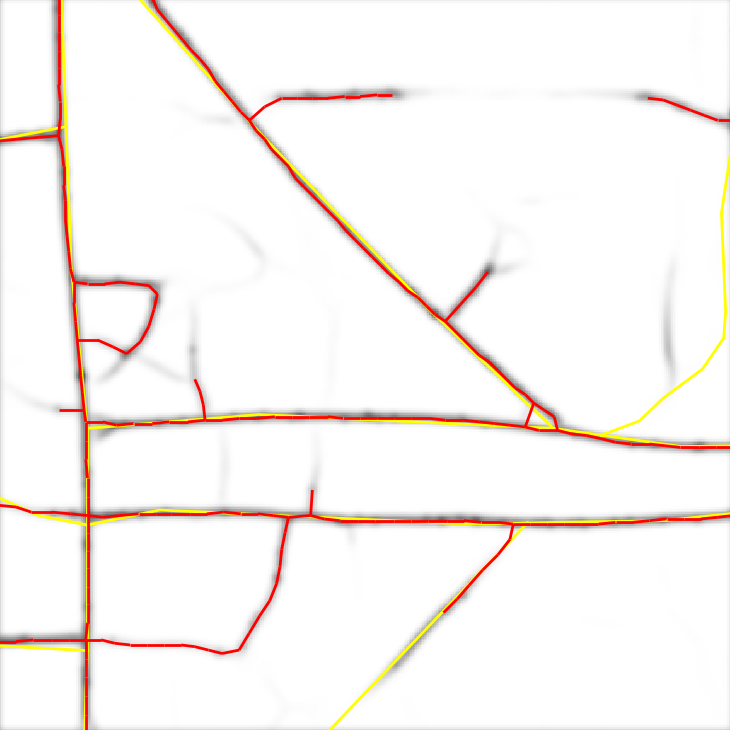}
        \caption*{$I_4^{te.}$ 0.6648 / 50.7188}
        \label{}
    \end{subfigure}
    \\
        \begin{subfigure}[b]{0.3\textwidth}
        \includegraphics[width=\textwidth]{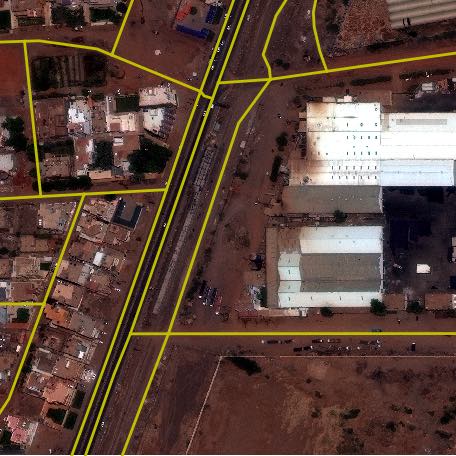}
        \caption*{AOI\_5 - Id: 150}
        \label{}
    \end{subfigure}
    \begin{subfigure}[b]{0.3\textwidth}
        \includegraphics[width=\textwidth]{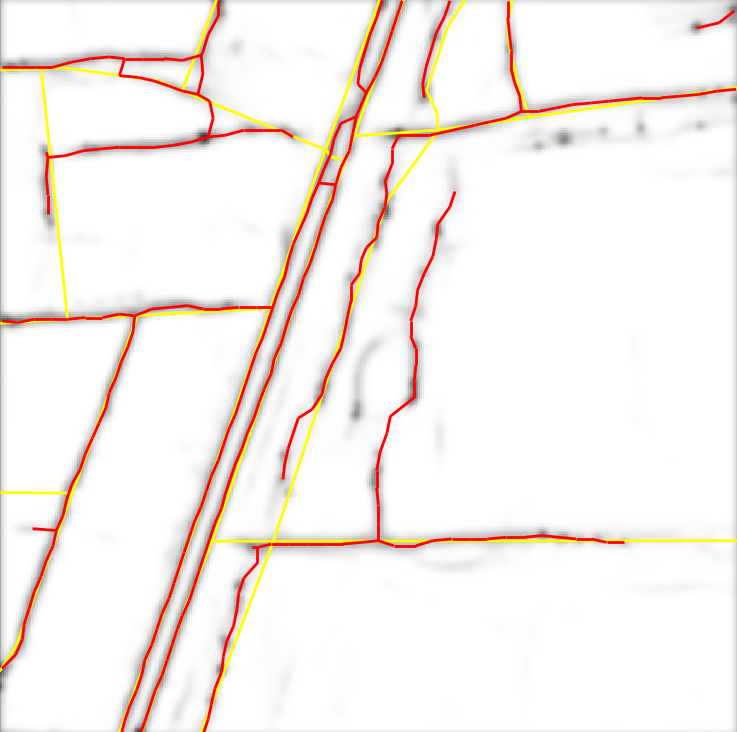}
        \caption*{$I_1^{te.}$ 0.3758 / 28.1091}
        \label{}
    \end{subfigure}
    \begin{subfigure}[b]{0.3\textwidth}
        \includegraphics[width=\textwidth]{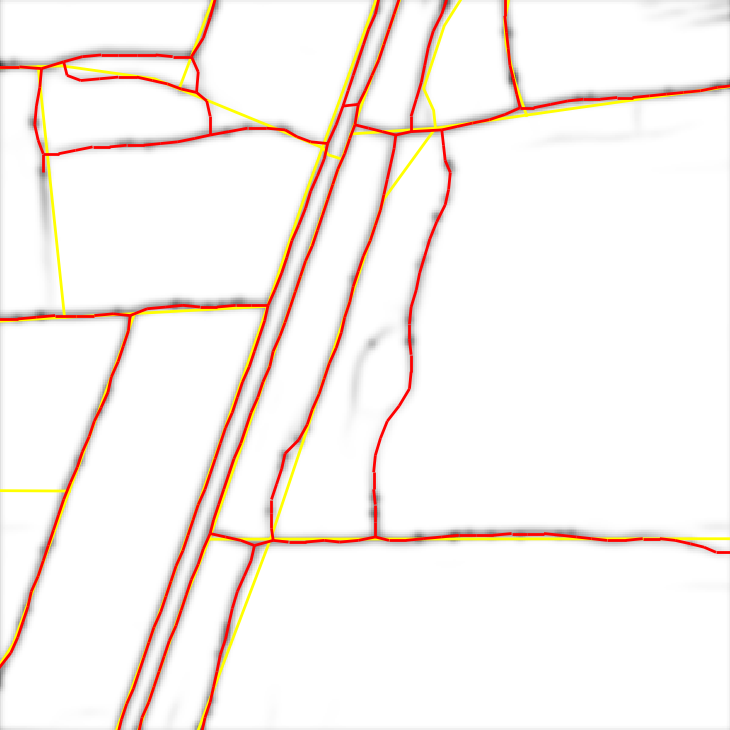}
        \caption*{$I_4^{te.}$ 0.7356 / 23.2734}
        \label{}
    \end{subfigure}
    \caption{Using 10\% labelled images. Reconstructed graphs by our \MorseTrain() after 1 and 4 iterations.}
    \label{fig:rand20_us}
\end{figure}

\paragraph{With 10\% ground truth}
Now we use a small set of labelled data: Specifically, we assume that only 10\% images (i.e, 20 images) have labels (i.e, ground-truth roads given). 
Table \ref{tab:10gt_APLS} shows the APLS-score for test images after different iterations by \MorseTrain() and \SkeTrain().
For \MorseTrain{}, all scores improve.
It is important to note that with only 10\% labeled-images, we can now also handle the challenging AOI\_5 dataset, and achieve an APLS-score of $0.607$. (For the case of AOI\_2, compared to the label-free case, the score of our new \MorseTrain() improves to $0.721$ from $0.592$). 

\begin{figure}[htbp]
    \centering
            \begin{subfigure}[b]{0.3\textwidth}
        \includegraphics[width=\textwidth]{fig/nogt/AOI2_429_ORI}
        \caption*{AOI\_2 429}
        \label{}
    \end{subfigure}
    \begin{subfigure}[b]{0.3\textwidth}
        \includegraphics[width=\textwidth]{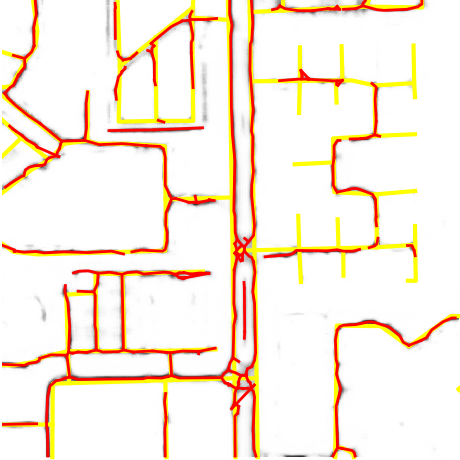}
        \caption*{$I_1^{te.}$ 0.2430 / 16.7267}
        \label{}
    \end{subfigure}
    \begin{subfigure}[b]{0.3\textwidth}
        \includegraphics[width=\textwidth]{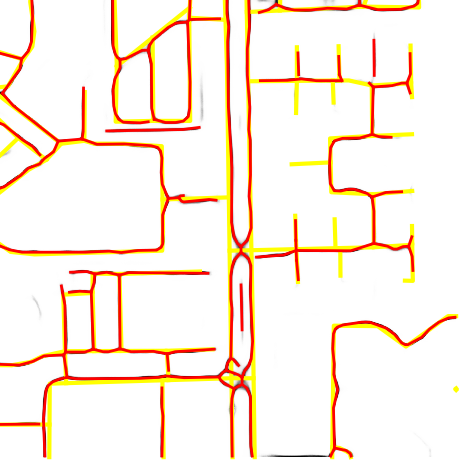}
        \caption*{$I_4^{te.}$ 0.2547 / 11.8099}
        \label{}
    \end{subfigure}
    \\
        \begin{subfigure}[b]{0.3\textwidth}
        \includegraphics[width=\textwidth]{fig/nogt/AOI5_50_ORI}
        \caption*{AOI\_5 50}
        \label{}
    \end{subfigure}
    \begin{subfigure}[b]{0.3\textwidth}
        \includegraphics[width=\textwidth]{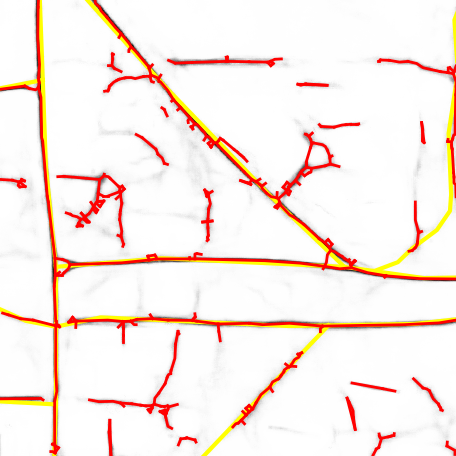}
        \caption*{$I_1^{te.}$ 0.4533 / 74.7638}
        \label{}
    \end{subfigure}
    \begin{subfigure}[b]{0.3\textwidth}
        \includegraphics[width=\textwidth]{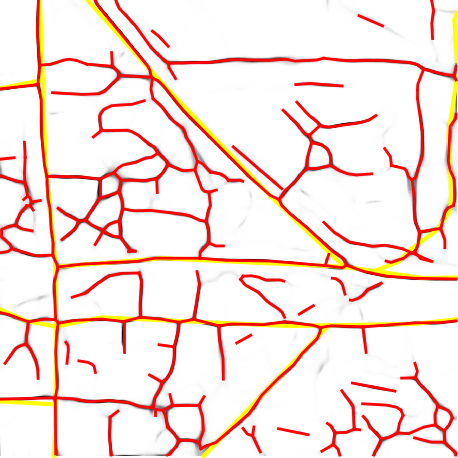}
        \caption*{$I_4^{te.}$ 0.1940 / 94.0461}
        \label{}
    \end{subfigure}
    \caption{Using 10\% labelled images. Reconstruction of the alternative \SkeTrain() method after 1 and 4 iterations.}
    \label{fig:rand20_ske}
\end{figure}

It is interesting to note that this iterative procedure does not seem to help  \SkeTrain() much, with scores even getting worse for the noisy dataset AOI\_5.
We show some examples of reconstructed graphs at different iterations for our algorithm (Figure \ref{fig:rand20_us}) and for the alternative \SkeTrain() method (Figure \ref{fig:rand20_ske}).

\subsection{Limitations and future work}
\label{sec:future_work}
\begin{figure}[htbp]
    \centering

    \begin{subfigure}[b]{0.3\textwidth}
        \includegraphics[width=\textwidth]{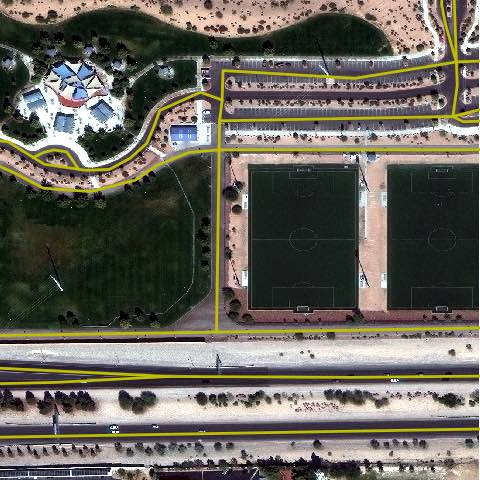}
        \caption*{AOI\_2-Id: 333}
        \label{}
    \end{subfigure}
    ~ 
    \begin{subfigure}[b]{0.3\textwidth}
        \includegraphics[width=\textwidth]{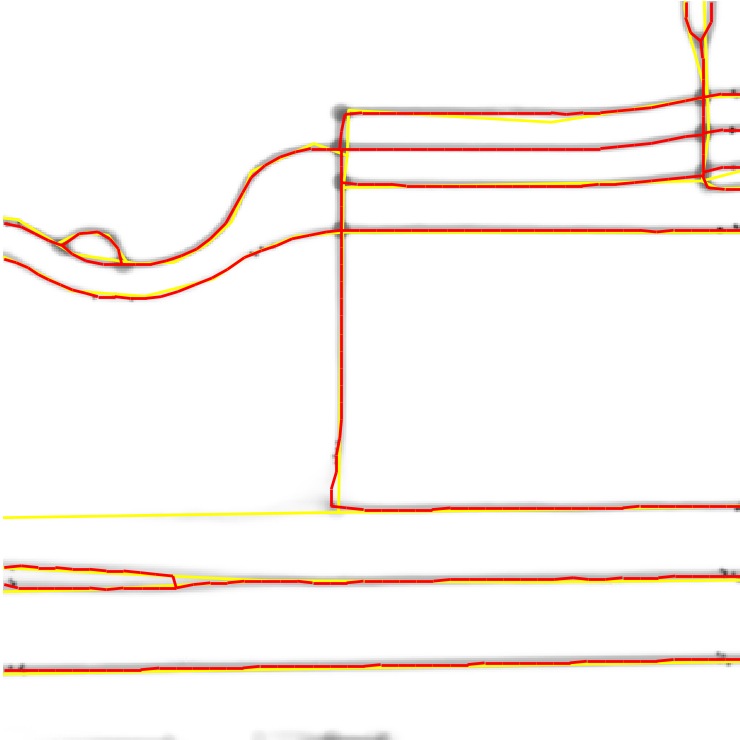}
        \caption*{$\delta$=0.1 / 0.8162 }
        \label{}
    \end{subfigure}
    ~
        \begin{subfigure}[b]{0.3\textwidth}
        \includegraphics[width=\textwidth]{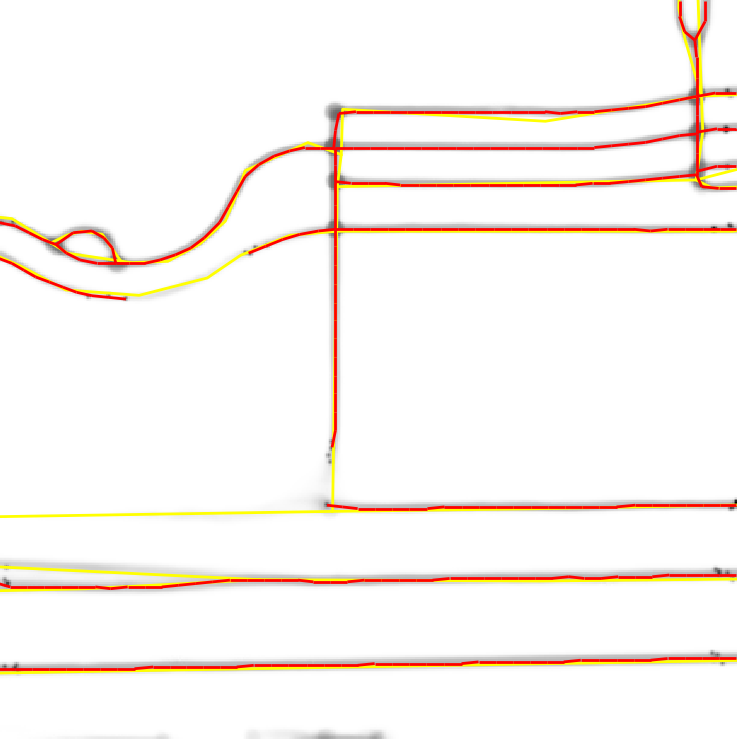}
        \caption*{$\delta$=0.15 / 0.6372  }
        \label{}
    \end{subfigure}\\
     \begin{subfigure}[b]{0.3\textwidth}
        \includegraphics[width=\textwidth]{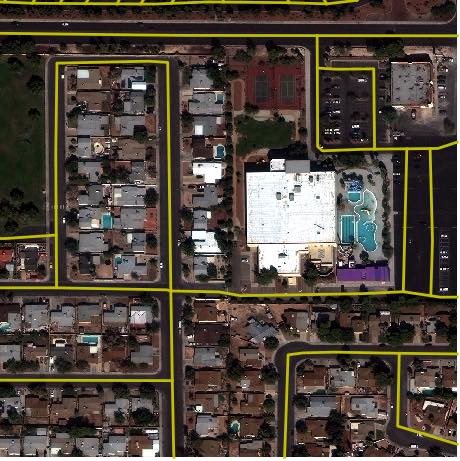}
        \caption*{AOI\_2-Id: 1107}
        \label{}
    \end{subfigure}
    ~ 
    \begin{subfigure}[b]{0.3\textwidth}
        \includegraphics[width=\textwidth]{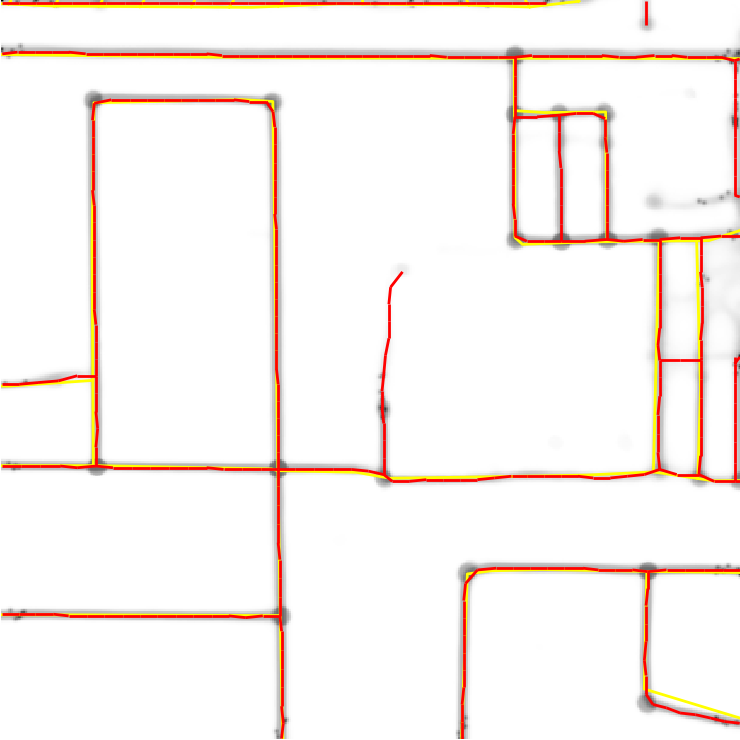}
        \caption*{$\delta$=0.1 / 0.8412 }
        \label{}
    \end{subfigure}
    ~
        \begin{subfigure}[b]{0.3\textwidth}
        \includegraphics[width=\textwidth]{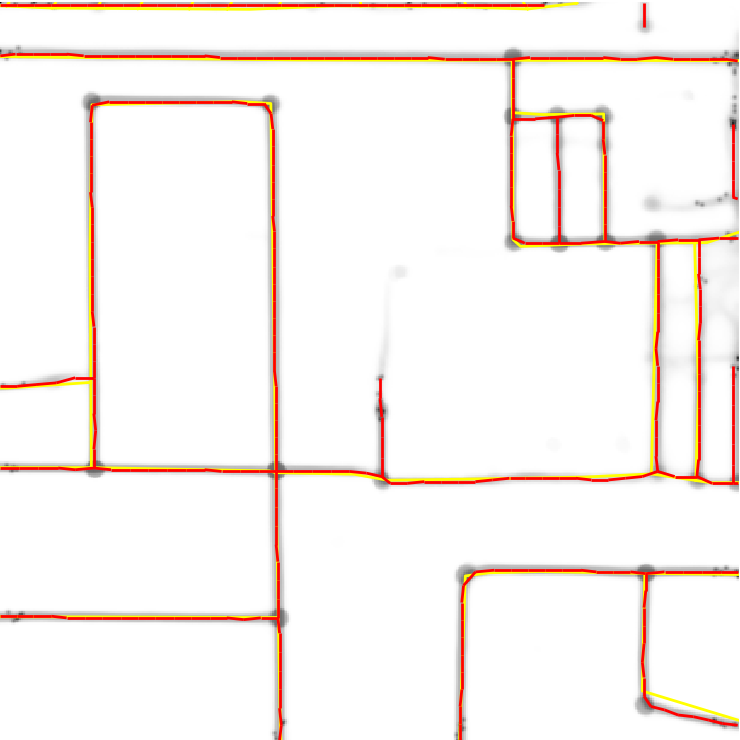}
        \caption*{$\delta$=0.15 / 0.9035 }
        \label{}
    \end{subfigure} 
    \caption{Effects of the persistence threshold $\delta$ on the results. Left: raw satellite images (yellow graphs are ground-truth road networks). Middle/right: Results for different  $\delta$, the number after $\delta$ is the APLS score. The first row gives an example where low $\delta$ leads to better results and the second row  gives an example where high $\delta$ leads to better results. \label{fig:para-delta}}
\end{figure}

\begin{figure}[htbp]
    \centering
 \begin{subfigure}[b]{0.3\textwidth}
        \includegraphics[width=\textwidth]{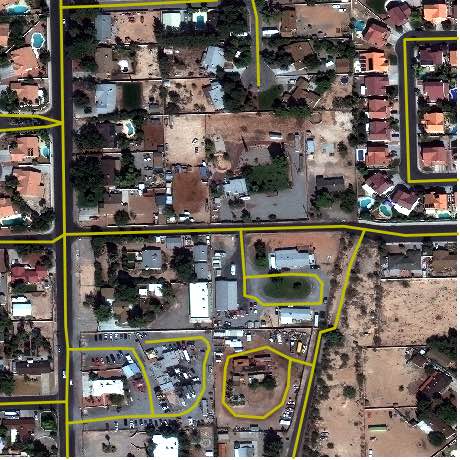}
        \caption*{AOI\_2-Id: 964}
        \label{}
    \end{subfigure}
    ~ 
    \begin{subfigure}[b]{0.3\textwidth}
        \includegraphics[width=\textwidth]{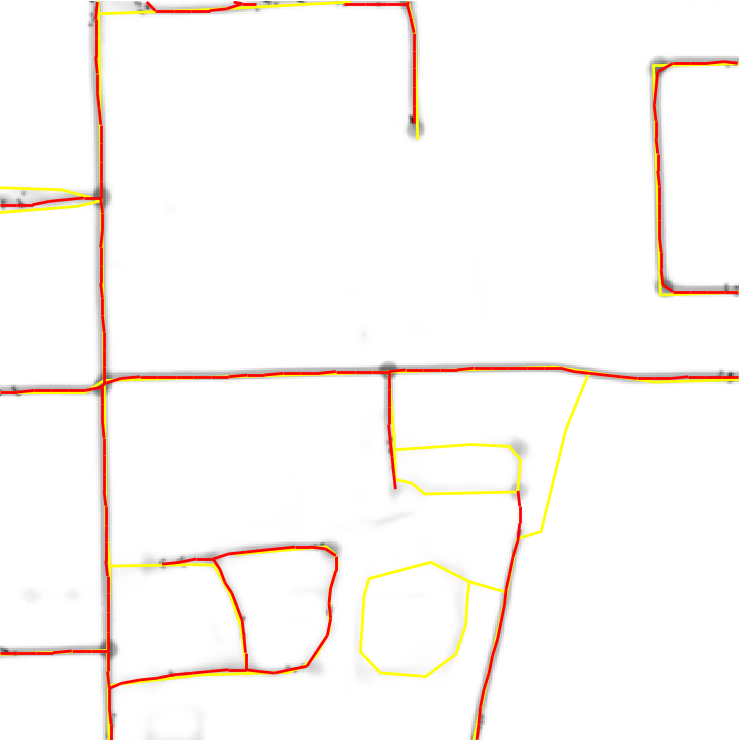}
        \caption*{$\tau$=0.4 / 0.5827 }
        \label{}
    \end{subfigure}
    ~
        \begin{subfigure}[b]{0.3\textwidth}
        \includegraphics[width=\textwidth]{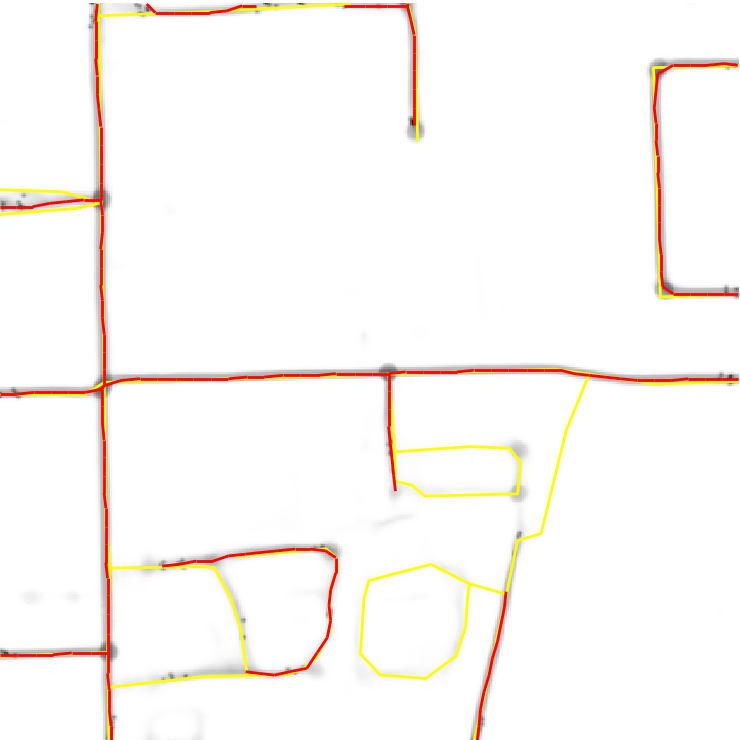}
        \caption*{$\tau$=0.5 / 0.4598 }
        \label{}
    \end{subfigure} \\
    \begin{subfigure}[b]{0.3\textwidth}
        \includegraphics[width=\textwidth]{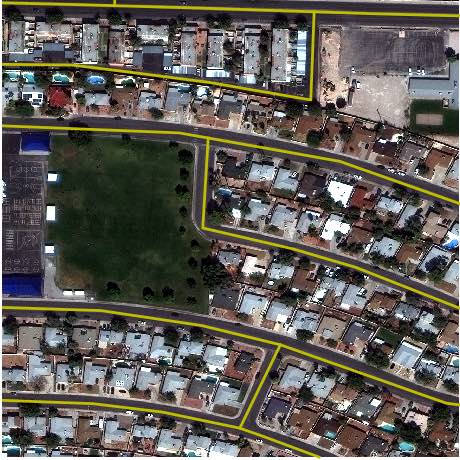}
        \caption*{AOI\_2-Id: 750}
        \label{}
    \end{subfigure}
    ~ 
    \begin{subfigure}[b]{0.3\textwidth}
        \includegraphics[width=\textwidth]{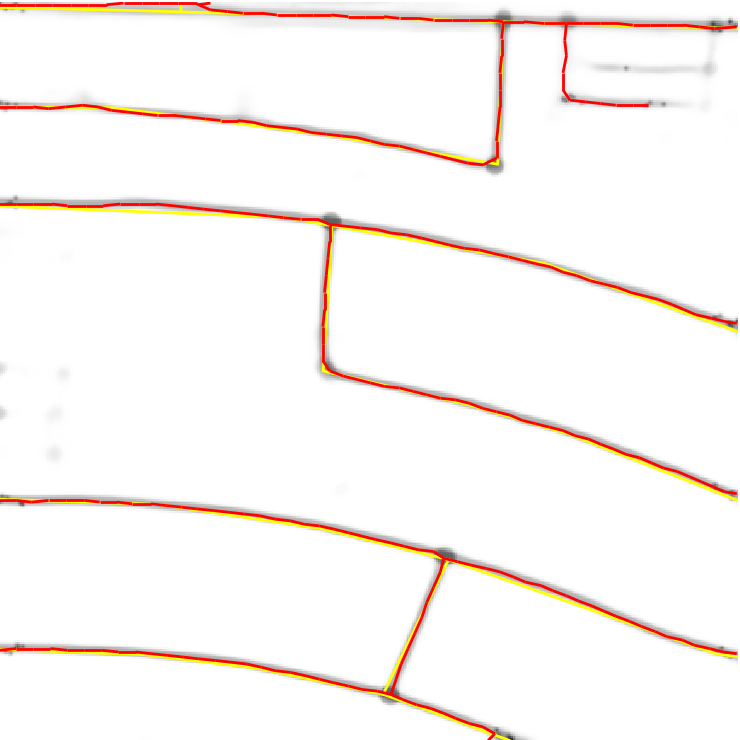}
        \caption*{$\tau$=0.4 / 0.9255 }
        \label{}
    \end{subfigure}
    ~
        \begin{subfigure}[b]{0.3\textwidth}
        \includegraphics[width=\textwidth]{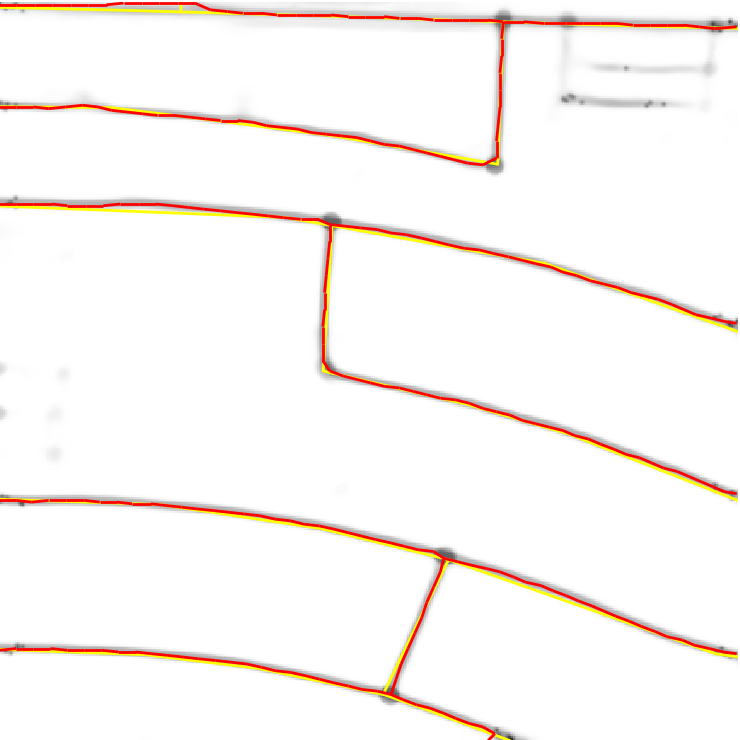}
        \caption*{$\tau$=0.5 / 0.9753  }
        \label{}
    \end{subfigure}
    \caption{Effects of the arc intensity threshold $\tau$. Left: raw satellite images (yellow graphs are ground-truth road networks). Middle/right: Results for different  $\tau$, the number after $\tau$ is the APLS score. The first row gives example when low $\tau$ leads to better results and the second row  gives example when high $\tau$ leads to better results. \label{fig:para-arc}}
\end{figure}

First, currently we choose the parameter $\delta$ globally.
Figure \ref{fig:para-delta} shows the effect of the persistence threshold $\delta$.
The example demonstrates that there is no single parameter value that works for all cases. 
As for the parameter arc-intensity threshold $\tau$,
we choose it adaptively for AOI\_3\_Shanghai and AOI\_4\_Paris by the intensity of the images to deal with the extreme sparse images.
For general cases, it is hard to make this choice simply based on the 
intensities of the images, see Figure \ref{fig:para-arc}.
An interesting future research direction would be to investigate how to choose 
these parameters adaptively, 
yet (semi-)automatically. 
Second, we recover the tips by locating their positions and modifying the density values.
It will be interesting to see if we can recover the tips from the 
graph reconstruction algorithm directly.
Third, we observe that the fully automatic framework sometimes is not efficient for a noisy dataset such as AOI\_5\_Khartoum.  It would be good to improve the performance of this approach for noisy datasets.


\vspace{0.1in}
\noindent
{\bf Acknowledgment}: We acknowledge the NSF grants CCF-1740761, RI-1815697, CCF-1733798 and CCF-1618247 for partially supporting this research.

\bibliography{references}

\begin{thebibliography}{10}

\bibitem{spacenetResult}
Winning solutions from spacenet road detection and routing challenge.
\newblock \url{https://github.com/SpaceNetChallenge/RoadDetector}, 2018.

\bibitem{albu}
A.~Buslaev.
\newblock Spacenet round 3 winner.
\newblock
  \url{https://github.com/SpaceNetChallenge/RoadDetector/tree/master/albu-solution},
  2018.

\bibitem{buslaev2018fully}
A.~Buslaev, S.~Seferbekov, V.~Iglovikov, and A.~Shvets.
\newblock Fully convolutional network for automatic road extraction from
  satellite imagery.
\newblock In {\em The IEEE Conference on Computer Vision and Pattern
  Recognition (CVPR) Workshops}, 2018.

\bibitem{ciresan2012deep}
D.~Ciresan, A.~Giusti, L.~M. Gambardella, and J.~Schmidhuber.
\newblock Deep neural networks segment neuronal membranes in electron
  microscopy images.
\newblock In {\em Advances in neural information processing systems}, pages
  2843--2851, 2012.

\bibitem{das2011use}
S.~Das, TT~Mirnalinee, and K.~Varghese.
\newblock Use of salient features for the design of a multistage framework to
  extract roads from high-resolution multispectral satellite images.
\newblock {\em IEEE transactions on Geoscience and Remote sensing},
  49(10):3906--3931, 2011.

\bibitem{DRS15}
O.~Delgado-Friedrichs, V.~Robins, and A.~Sheppard.
\newblock Skeletonization and partitioning of digital images using discrete
  morse theory.
\newblock {\em IEEE Trans. Pattern Anal. Machine Intelligence}, 37(3):654--666,
  March 2015.

\bibitem{dey2017improved}
T.~K. Dey, J.~Wang, and Y.~Wang.
\newblock Improved road network reconstruction using discrete morse theory.
\newblock In {\em Proc. 25th ACM SIGSPATIAL}, page~58. ACM, 2017.

\bibitem{dey2018graph}
T.~K. Dey, J.~Wang, and Y.~Wang.
\newblock {Graph Reconstruction by Discrete Morse Theory}.
\newblock In {\em Proc. 34th Sympos. Comput. Geom.}, pages 31:1--31:15, 2018.

\bibitem{EH10}
H.~Edelsbrunner and J.~Harer.
\newblock {\em Computational Topology : an Introduction}.
\newblock American Mathematical Society, 2010.

\bibitem{ELZ02}
H.~Edelsbrunner, D.~Letscher, and A.~Zomorodian.
\newblock Topological persistence and simplification.
\newblock {\em Discr. Comput. Geom.}, 28:511--533, 2002.

\bibitem{forman}
R.~Forman.
\newblock Morse theory for cell complexes.
\newblock {\em Advances in mathematics}, 134(1):90--145, 1998.

\bibitem{gu2015fusion}
X.~Gu, A.~Zang, X.~Huang, A.~Tokuta, and X.~Chen.
\newblock Fusion of color images and lidar data for lane classification.
\newblock In {\em Proc. 23rd ACM SIGSPATIAL}, page~69. ACM, 2015.

\bibitem{GDN07}
A.~Gyulassy, M.~Duchaineau, V.~Natarajan, V.~Pascucci, E.~Bringa,
  A.~Higginbotham, and B.~Hamann.
\newblock Topologically clean distance fields.
\newblock {\em IEEE Trans. Visualization Computer Graphics}, 13(6):1432--1439,
  Nov 2007.

\bibitem{he2016deep}
K.~He, X.~Zhang, S.~Ren, and J.~Sun.
\newblock Deep residual learning for image recognition.
\newblock In {\em Proc. of the IEEE conference on computer vision and pattern
  recognition}, pages 770--778, 2016.

\bibitem{RWS11}
V.~Robins, P.~J. Wood, and A.~P. Sheppard.
\newblock Theory and algorithms for constructing discrete morse complexes from
  grayscale digital images.
\newblock {\em IEEE Trans. Pattern Anal. Machine Intelligence},
  33(8):1646--1658, Aug 2011.

\bibitem{ronneberger2015u}
O.~Ronneberger, P.~Fischer, and T.~Brox.
\newblock U-net: Convolutional networks for biomedical image segmentation.
\newblock In {\em International Conference on Medical image computing and
  computer-assisted intervention}, pages 234--241. Springer, 2015.

\bibitem{shi2014integrated}
W.~Shi, Z.~Miao, and J.~Debayle.
\newblock An integrated method for urban main-road centerline extraction from
  optical remotely sensed imagery.
\newblock {\em IEEE Transactions on Geoscience and Remote Sensing},
  52(6):3359--3372, 2014.

\bibitem{sknw}
sknw.
\newblock \url{https://github.com/yxdragon/sknw}, 2017.

\bibitem{2011MNRAS}
T.~{Sousbie}.
\newblock {The persistent cosmic web and its filamentary structure - I. Theory
  and implementation}.
\newblock 414:350--383, June 2011.
\newblock \href {http://arxiv.org/abs/1009.4015} {\path{arXiv:1009.4015}}.

\bibitem{sun2018combining}
T.~Sun, Z.~Di, and Y.~Wang.
\newblock Combining satellite imagery and gps data for road extraction.
\newblock In {\em Proc. of the 2nd ACM SIGSPATIAL International Workshop on AI
  for Geographic Knowledge Discovery}, pages 29--32. ACM, 2018.

\bibitem{van2018spacenet}
A.~Van~Etten, D.~Lindenbaum, and T.~M. Bacastow.
\newblock Spacenet: A remote sensing dataset and challenge series.
\newblock {\em arXiv preprint arXiv:1807.01232}, 2018.

\bibitem{wang2015efficient}
S.~Wang, Y.~Wang, and Y.~Li.
\newblock Efficient map reconstruction and augmentation via topological
  methods.
\newblock In {\em Proc. 23rd ACM SIGSPATIAL}, page~25. ACM, 2015.

\bibitem{weiss2013primal}
K.~Weiss, F.~Iuricich, R.~Fellegara, and L.~De~Floriani.
\newblock A primal/dual representation for discrete morse complexes on
  tetrahedral meshes.
\newblock In {\em Computer Graphics Forum}, volume~32, pages 361--370, 2013.

\bibitem{zang2017lane}
A.~Zang, R.~Xu, Z.~Li, and D.~Doria.
\newblock Lane boundary extraction from satellite imagery.
\newblock In {\em Proc. of the 1st ACM SIGSPATIAL Workshop on High-Precision
  Maps and Intelligent Applications for Autonomous Vehicles}, page~1. ACM,
  2017.

\bibitem{zhao2017pyramid}
H.~Zhao, J.~Shi, X.~Qi, X.~Wang, and J.~Jia.
\newblock Pyramid scene parsing network.
\newblock In {\em IEEE Conf. on Computer Vision and Pattern Recognition
  (CVPR)}, pages 2881--2890, 2017.

\end{thebibliography}

%
%
%
%
%
%
%
%
%
\end{document}